\documentclass[10pt,twocolumn,letterpaper]{article}

\usepackage[pagenumbers]{cvpr} 

\usepackage{graphicx}
\usepackage{amsmath}
\usepackage{amssymb}
\usepackage{booktabs}
\usepackage[english]{babel} 
\usepackage[T1]{fontenc}
\usepackage{array}
\usepackage{multirow}
\usepackage{amstext}
\usepackage{bbding}
\usepackage[accsupp]{axessibility}

\makeatletter

\providecommand{\tabularnewline}{\\}

\makeatother

\usepackage{babel}

\usepackage[pagebackref,breaklinks,colorlinks]{hyperref}

\usepackage[capitalize]{cleveref}
\crefname{section}{Sec.}{Secs.}
\Crefname{section}{Section}{Sections}
\Crefname{table}{Table}{Tables}
\crefname{table}{Tab.}{Tabs.}

\def\cvprPaperID{3915} 
\def\confName{CVPR}
\def\confYear{2022}

\begin{document}

\title{Incremental Transformer Structure Enhanced Image Inpainting with Masking Positional Encoding}

\author{Qiaole Dong\footnotemark[1], Chenjie Cao\footnotemark[1], Yanwei Fu\footnotemark[2]\\
School of Data Science, Fudan University\\
{\tt\small \{18307130096,20110980001,yanweifu\}@fudan.edu.cn}
}
\maketitle

\renewcommand{\thefootnote}{\fnsymbol{footnote}} 
\footnotetext[1]{Equal contributions.}
\footnotetext[2]{Corresponding authors.}

\begin{abstract}
Image inpainting has made significant advances in recent years. However, it is still challenging to recover corrupted images with both vivid textures and reasonable structures. Some specific methods only tackle regular textures while losing holistic structures due to the limited receptive fields of convolutional neural networks (CNNs). On the other hand, attention-based models can learn better long-range dependency for the structure recovery, but they are limited by the heavy computation for inference with large image sizes. To address these issues, we propose to leverage an additional structure restorer to facilitate the image inpainting incrementally. The proposed model restores holistic image structures with a powerful attention-based transformer model in a fixed low-resolution sketch space. Such a grayscale space is easy to be upsampled to larger scales to convey correct structural information. Our structure restorer can be integrated with other pretrained inpainting models efficiently with the zero-initialized residual addition. Furthermore, a masking positional encoding strategy is utilized to improve the performance with large irregular masks. Extensive experiments on various datasets validate the efficacy of our model compared with other competitors. Our codes are released in \url{https://github.com/DQiaole/ZITS_inpainting}.
\end{abstract}

 \vspace{-0.1in}
\section{Introduction}
\label{sec:intro}

\begin{figure}
\begin{centering}
\includegraphics[width=0.99\linewidth]{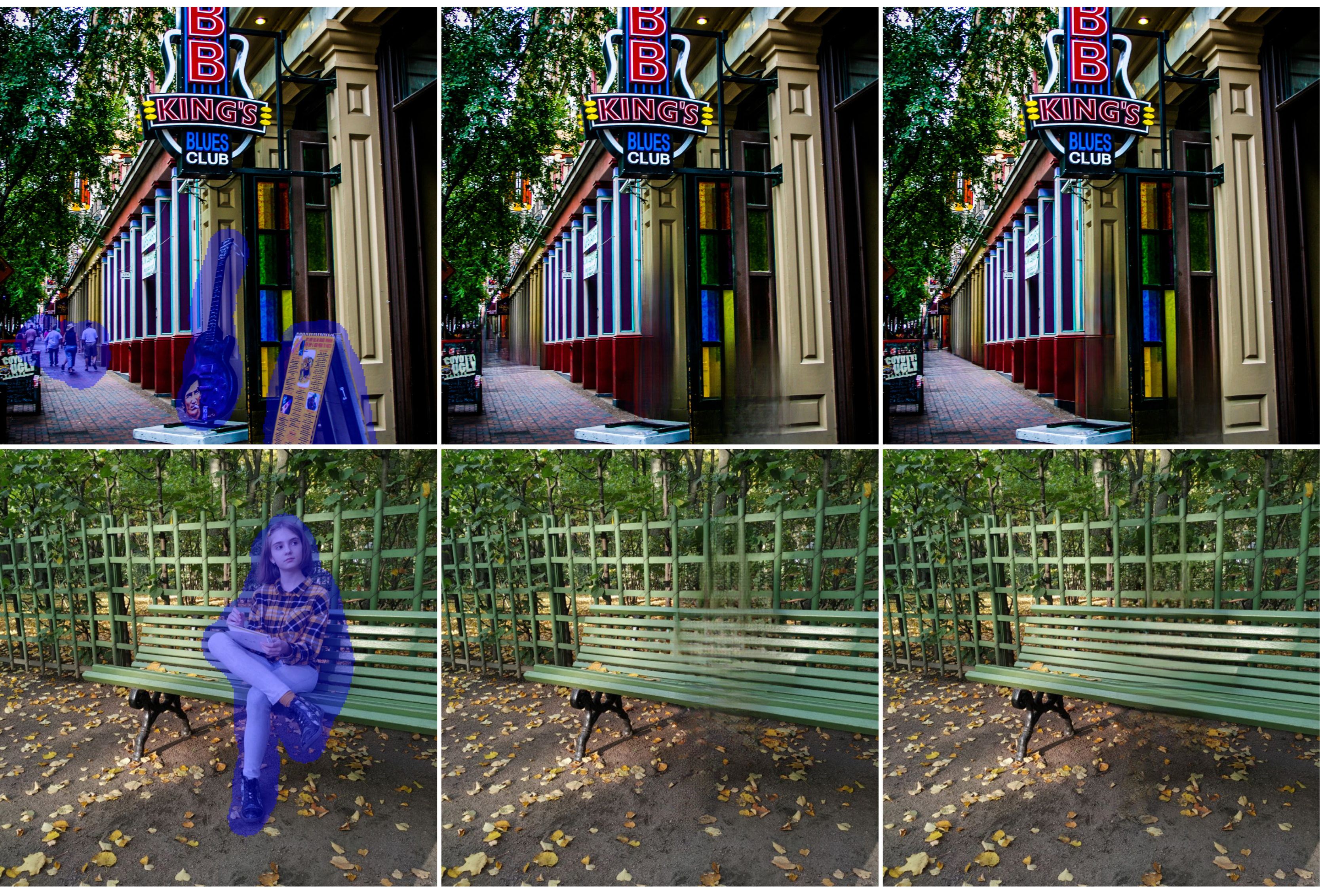}
\par\end{centering}
 \vspace{-0.1in}
 \caption{High quality 1024$\times$1024 inpainted results. From left to right, masked inputs, results of LaMa~\cite{suvorov2021resolution}, results of our method.
 \label{fig:teaser}}
 \vspace{-0.15in}
\end{figure}

Image inpainting has been investigated as a long-standing challenge to address the difficulty of filling in missing areas of pictures.
It is very useful to various real-world applications, such as object removal~\cite{elharrouss2020image}, photo restoration, and image editing~\cite{jo2019sc}.
To achieve realistic outcomes, the inpainted images should remain both semantically coherent textures and visually reasonable structures. Many classical algorithms~\cite{Bertalmo2000ImageI,Levin2003LearningHT,Roth2005FieldsOE,Hays:2007,criminisi2003object} search similar patches for the reconstruction heuristically.
But preserving good textures and holistic structures in large images is still non-trivial for these conventional methods.

Benefited from excellent capacities of Convolutional Neural Networks (CNNs)~\cite{krizhevsky2012imagenet} and Generative Adversarial Networks (GANs)~\cite{goodfellow2014generative}, existing deep learning  methods~\cite{liu2018image,zeng2019learning, lahiri2020prior,guo2021image,cao2021learning,wan2021highfidelity,yi2020contextual,suvorov2021resolution} 
could  efficiently conduct the image inpainting tasks in some common cases. 
However, they still suffer from some dilemmas. \emph{(1) Limited receptive fields.} Learning semantically consistent textures is difficult for traditional CNNs due to the local inductive priors and narrow receptive field of convolution operations. Even dilated convolutions~\cite{yu2015multi} fail to tackle large corrupted regions or high-resolution images. \emph{(2) Missing holistic structures.} Recovering key edges and lines for scenes, especially ones with weak texture is difficult without the holistic understanding of large images as shown in Fig.~\ref{fig:teaser}. \emph{(3) Heavy computations.} Training GANs with large image sizes is still very tricky and costly~\cite{karras2020analyzing}. And the inpainting performance may be degraded on high-resolution images.
\emph{(4) No positional information in masked regions.} The inpainting model tends to repeat meaningless artifacts in large irregular masked regions without explicit positional clues.

Some pioneering works can partially solve these problems. 
For the limited receptive fields, attention-based methods~\cite{yu2018generative,yi2020contextual,zeng2020high} leverage the attention mechanism to extend the receptive fields. Suvorov~\etal~\cite{suvorov2021resolution} utilize the Fast Fourier Convolution (FFC) to encode features in frequency fields with global receptive fields for resolution-robust inpainting.
But they fail to ensure the holistic structures and work inferior for the images of weak texture.
Furthermore, transformer-based  methods~\cite{wan2021highfidelity,yu2021diverse} with long-range dependency are utilized to firstly fill low-resolution tokens, and then upsample them with CNNs. Unfortunately, transformers demand huge memory footprint for large images. And the resolution disparity between transformer and CNN causes serious error propagation.
On the other hand, some methods utilize auxiliary information for structure recovery, \emph{e.g.}, edges~\cite{nazeri2019edgeconnect,guo2021image}, segmentation~\cite{liao2020guidance,song2018spg}, and gradients~\cite{yang2020learning}. Cao~\etal~\cite{cao2021learning} propose a sketch tensor space consisting of edges and wireframes~\cite{huang2018learning} to facilitate holistic structure learning in man-made scenes.
However, these sophisticated methods are usually based on multi-stage or multi-model designs, which are costly to be trained from scratch.
Moreover, many researches~\cite{islam2020much,xu2020positional,lin2021infinitygan} show that the position information is critical to learning the network, such as GANs~\cite{xu2020positional, lin2021infinitygan} and NeRF~\cite{NeRF}. To our knowledge, there is no previous  work that has explicitly discussed and utilized the position information in image inpainting.

Therefore, this motivates our work of 
incrementally inferring the holistic structural information and positional information to boost the performance of the inpainting model. Specifically, we leverage a transformer-based model to tackle holistic structures with edges and lines as the sketch tensor space~\cite{cao2021learning}. Critically, such a normalized grayscale space can be easily upsampled by a simple CNN to higher resolutions without information loss. Further, we propose a novel incrementally training strategy with Zero-initialized Residual Addition (ZeroRA)~\cite{bachlechner2020rezero} to incorporate the structural information into a pretrained inpainting model. 
This incremental strategy enjoys fast convergence for much fewer steps compared with retraining a new auxiliary-based model. Furthermore, we introduce the positional encoding for the mask region, which improves the performance of image restoration.

Formally, this paper proposes a novel ZeroRA based Incremental Transformer Structure  (ZITS) inpainting framework enhanced with Masking Positional Encoding (MPE). Our ZITS has novel components of Masking Positional Encoding (MPE), Transformer Structure Restorer (TSR), Fourier CNN Texture Restoration (FTR), and Structure Feature Encoder (SFE). 
The TSR is composed of alternating axial~\cite{ho2019axial} and standard attention blocks for the balance between the performance and the efficiency. Note that our TSR can achieve much better structure recovery compared with CNNs~\cite{nazeri2019edgeconnect,cao2021learning}. The output grayscale edges and lines are upsampled with a simple 4-layer CNN. Then, a gated convolutions~\cite{yu2019free} based SFE encodes features and transfers them to a FFC based inpainting model called FTR with ZeroRA. Furthermore, we use MPE to express both distances and directions from unmasked regions to masked ones.

We highlight several contributions as follows. (1) We propose using a transformer to learn a normalized grayscale sketch tensor space for inpainting tasks. Such an attention-based model can learn substantially better holistic structures with long-range dependency. (2) The auxiliary information can be incrementally incorporated into a pretrained inpainting model without retraining. (3) A novel masking positional encoding is proposed to improve the generalization of the inpainting model for different masks. (4) Extensive experiments on several datasets, which include Places2~\cite{zhou2017places}, ShanghaiTech~\cite{huang2018learning}, NYUDepthV2~\cite{Silberman:ECCV12}, and MatterPort3D~\cite{chang2017matterport3d} reveal that our proposed model outperforms other state-of-the-art competitors. 

\section{Related Work}

\noindent \textbf{Inpainting by Auxiliaries}.
Auxiliary information such as edges~\cite{nazeri2019edgeconnect,yang2020learning}, segmentation maps~\cite{song2018spg, liao2020guidance}, and gradients~\cite{yang2020learning} are shown very useful to inpainting. Specifically, EdgeConnect~\cite{nazeri2019edgeconnect} utilizes edges to help inpainting images with certain structures. Guo~\etal~\cite{guo2021image} propose a two-stream network for image inpainting, which models the structure constrained texture synthesis and texture-guided structure reconstruction in a coupled manner. SGE-Net~\cite{liao2020guidance} just iteratively updates the semantic segmentation maps and the corrupted image. Cao~\etal~\cite{cao2021learning} further propose learning a sketch tensor space, composed of edges and lines for inpainting man-made scenes. In our work, we also take edges and lines as our auxiliary information. Differs from~\cite{cao2021learning}, the transformer is leveraged to rebuild edges and lines in ZITS. As some preliminary investigations~\cite{chen2020generative} have shown its excellent capability in modeling structural relationships for natural image synthesis. Besides, almost all auxiliary-based methods need extra input channels for more information, which makes them must be retrained from scratch to utilize these additional inputs sufficiently. In our paper, we propose a flexible and effective way to add structural information to a pretrained inpainting model incrementally.

\noindent \textbf{Transformers for Image Generation}.
Transformer~\cite{vaswani2017attention,ba2016layer} achieved good performance on many tasks in the NLP and CV communities, as it learned long-range interactions on sequential data. 
Dosovitskiy~\etal~\cite{dosovitskiy2020image} firstly propose the use of a transformer for image recognition and show its great capacity. 
Many works~\cite{peters2018deep,kumar2021colorization,chen2020generative} devote to reducing the time and space complexity for transformers.
Esser~\etal~\cite{esser2021taming} and Ramesh~\etal~\cite{ramesh2021zeroshot} leverage discrete representation learning for lower computational cost. The transformer is also used in image inpainting~\cite{wan2021highfidelity,yu2021diverse} for low-resolution images reconstruction, and then guides the GAN-based CNN for further high-quality results. In our work, the transformer is used to build the holistic structure reconstruction and then to guide the image inpainting, which enjoys excellent performance compared with CNN-based methods.


\section{Method}
\begin{figure*}
\begin{centering}
\includegraphics[width=0.85\linewidth]{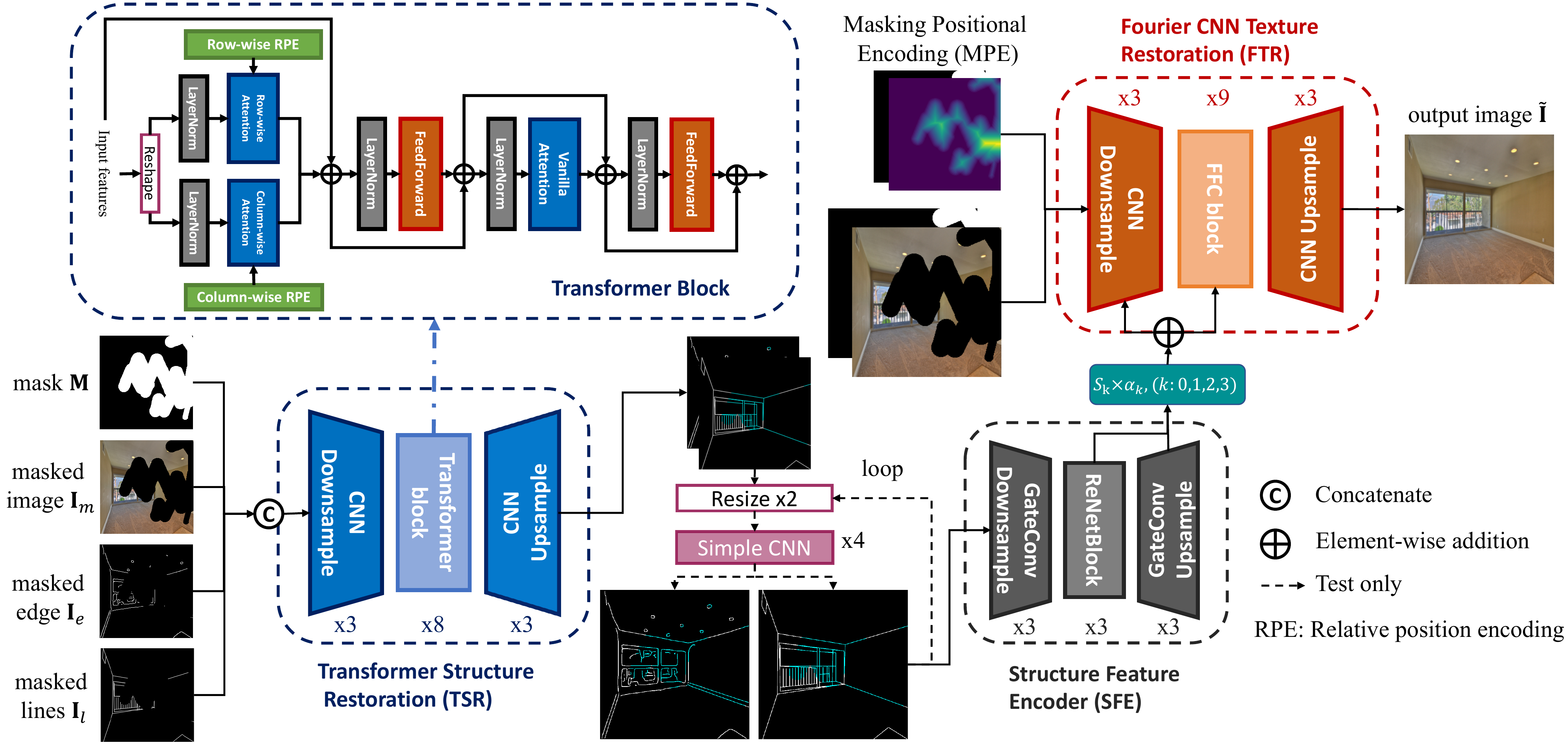}
\par\end{centering}
\vspace{-0.1in}
 \caption{The overview of our ZITS. At first, the TSR model is used to restore structures with low resolutions. Then the simple CNN based upsampler is leveraged to upsample edge and line maps. Moreover, the upsampled sketch space is encoded by the SFE model, and added to the FTR through ZeroRA to restore the textures. The top left corner show details about the transformer block. The input feature are learned through row-wise and column-wise attentions respectively, then encoded by a standard attention module.\label{fig:overview}}

\vspace{-0.1in}
\end{figure*}


\noindent\textbf{Overview.} The whole pipeline of ZITS is shown in Fig.~\ref{fig:overview}. Given masked image $\textbf{I}_m$, canny edge $\textbf{I}_{e}$~\cite{ding2001canny}, lines $\textbf{I}_{l}$~\cite{cao2021learning}, and binary mask $\textbf{M}$, we concatenate and input them to the Transformer Structure Restoration (TSR) model for recovered edges and lines as sketch space $[\tilde{\textbf{I}}_{e},\tilde{\textbf{I}}_{l}]=\mathrm{TSR}(\textbf{I}_m,\textbf{I}_{e},\textbf{I}_{l},\textbf{M})$ (Sec.~\ref{sec:TSR}). During the inference stage, the Simple Structure Upsampler (SSU) can easily upsample the grayscale sketch maps into arbitrary sizes (Sec.~\ref{sec:SSU}). 
Then, a gated convolution based Structure Feature Encoder (SFE) extracts multi-scale features $\mathbf{S}_k=\mathrm{SFE}(\tilde{\textbf{I}}_{e},\tilde{\textbf{I}}_{l},\textbf{M}), \{k=0,1,2,3\}$ from upsampled sketches. We incrementally add $\mathbf{S}_k$ to related layers of the Fourier convolution based CNN Texture Restoration (FTR)  as $\tilde{\textbf{I}}=\mathrm{FTR}(\textbf{I}_m,\textbf{M},\alpha_k\cdot\textbf{S}_k), \{k=0,1,2,3\}$ with the residual addition of zero-initialized trainable coefficients $\alpha_k$, \emph{i.e.}, ZeroRA (Sec.~\ref{sec:SFEandFTR}).

\subsection{Transformer Structure Restoration}
\label{sec:TSR}

Since the transformer shows an ability to get expressive global structure recoveries~\cite{wan2021highfidelity}, we leverage the capacity of the transformer to restore holistic structures in a relatively low resolution.
For the input masked image $\textbf{I}_m$, edges $\textbf{I}_e$, lines $\textbf{I}_l$, and mask $\textbf{M}$ in $256\times256$, we firstly downsample them with three convolutions to reduce computation for the attention learning. Such simple convolutions can also inject beneficial convolutional inductive bias for vision transformers compared with the patch based MLP embedding~\cite{xiao2021early}.
Then we add a learnable absolute position embedding to the feature at each spatial position and get $\mathbf{X}\in\mathbb{R}^{h\times w\times c}$ for the input to attention layers, where $h,w=32$ are height and width, and $c=256$ is the feature channel.

To overcome the quadratic complexity of the standard self-attention~\cite{vaswani2017attention}, we alternately use axial attention modules~\cite{ho2019axial} and standard attention modules as shown in the top left of Fig.~\ref{fig:overview}. The axial attention module can be implemented easily by adjusting the tensor shape for row-wise and column-wise and then processing them with dot product-based self-attention respectively.
To improve the spatial relation, we also provide relative position encoding (RPE)~\cite{raffel2019exploring} for each axial-attention module. For the input feature $\mathbf{X}\in\mathbb{R}^{h\times w\times c}$, we suppose that $\mathbf{x}_{ri,rj}, \mathbf{x}_{ci,cj}\in\mathbb{R}^{c}$ mean feature vectors of rows $i,j$ and columns $i,j$ of $\mathbf{X}$.
Then the row and column-wise RPE based axial attention scores $\mathbf{A}^{row}, \mathbf{A}^{col}$ can be written as
\vspace{-0.05in}
\begin{equation}
\vspace{-0.05in}
\begin{split}
\mathbf{A}_{i,j}^{row}&=\mathbf{x}_{ri}\mathbf{W}_{rq}\mathbf{W}_{rk}^T\mathbf{x}_{rj}^T+R_{i,j}^{row},\\
\mathbf{A}_{i,j}^{col}&=\mathbf{x}_{ci}\mathbf{W}_{cq}\mathbf{W}_{ck}^T\mathbf{x}_{cj}^T+R_{i,j}^{col},
\end{split}
\label{eq:axial_attention}
\end{equation}
where $\mathbf{W}_{rq},\mathbf{W}_{rk},\mathbf{W}_{cq},\mathbf{W}_{ck}$ are trainable parameters for query and key in row and column; $R_{i,j}^{row}$ is the trainable RPE value between row $i$ and $j$, and $R_{i,j}^{col}$ means the RPE value between columns $i,j$. Then, the attention scores are processed by the softmax operation.
To stabilize the training, we use the pre-norm trick in~\cite{xiong2020layer}.
Compared with the $O(n^2)$ complexity of the standard self-attention, the axial attention only has $O(2n^{\frac{3}{2}})$, which allows us can handle more attention layers for a better capacity. 
Besides, we also remain some standard attention modules for learning the global correlation.
Our ablation shows that this setting can improve the performance with the same memory cost.

After the encoding of stacked transformer blocks, features are upsampled by three transpose convolutions for outputting structures in 256$\times$256. We use the binary cross-entropy (BCE) loss to optimize the predicted continuous sketch structures of edges $\mathbf{\tilde{I}}_e$ and lines $\mathbf{\tilde{I}}_l$ as
\vspace{-0.05in}
\begin{equation}
\vspace{-0.05in}
\mathcal{L}_{e}=\mathrm{BCE}(\mathbf{\tilde{I}}_e, \mathbf{\hat{I}}_e),\;\;\mathcal{L}_{l}=\mathrm{BCE}(\mathbf{\tilde{I}}_l, \mathbf{\hat{I}}_l),
\label{eq:tsr_loss}
\end{equation}
where $\mathbf{\hat{I}}_e$ means the binary ground truth canny edges, and $\mathbf{\hat{I}}_l$ indicates the antialiasing lines map got from the masking augmented wireframe detector from~\cite{cao2021learning}.



\subsection{Simple Structure Upsampler}
\label{sec:SSU}

\begin{figure}
\begin{centering}
\includegraphics[width=0.85\linewidth]{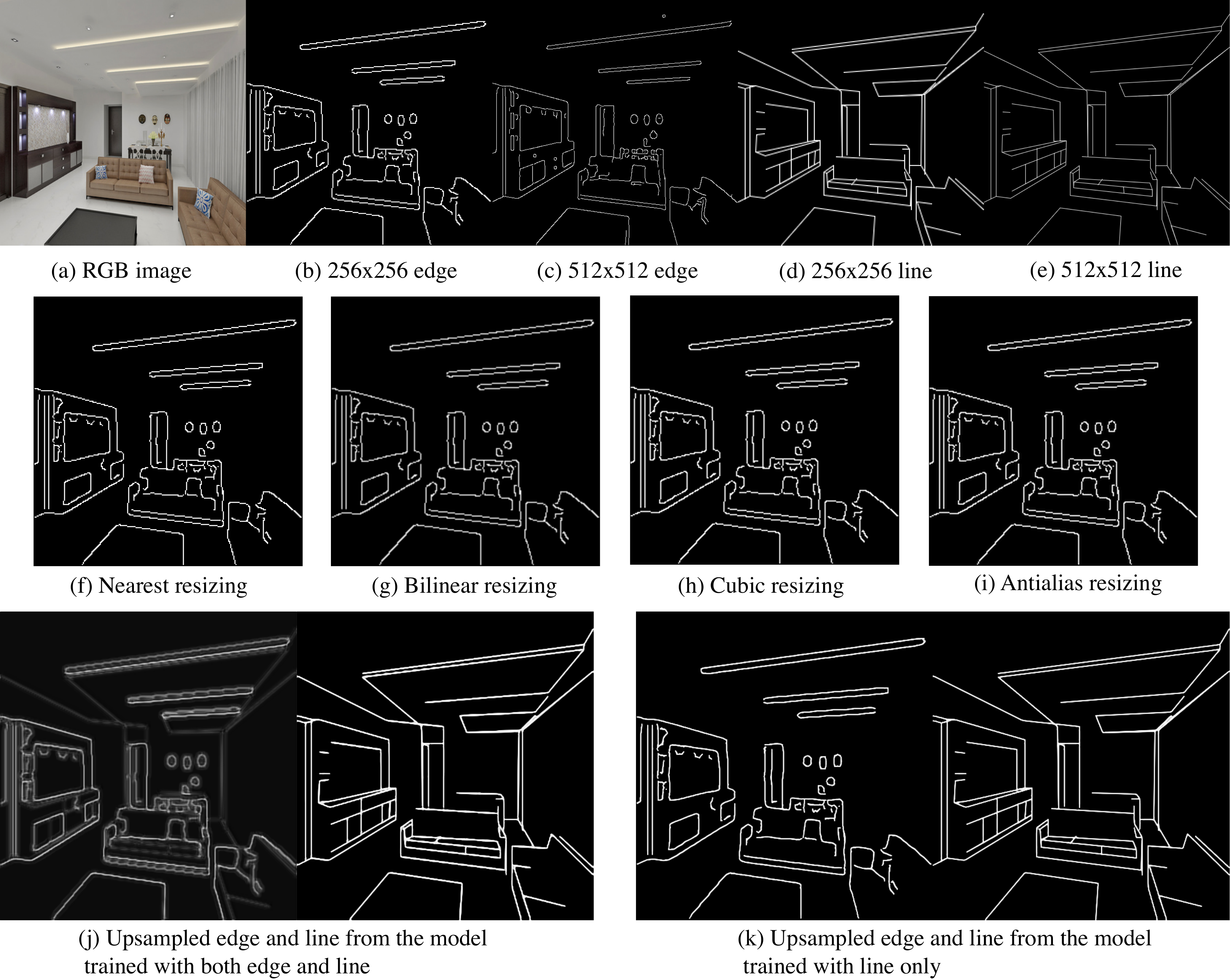}
\par\end{centering}
\vspace{-0.1in}
 \caption{(a)--(e) indicate the ground truth images and structures. Edges are got from canny edge detector, while sigma is 2.0 for 256$\times$256 and 2.5 for 512$\times$512. However, there are obvious ambiguities between (b) and (c). (f)--(i) show resizing edges of different interpolations. The learning based upsampling edge from (j) has significant superior quality compared with one from (k). \label{fig:upsample}}
\vspace{-0.15in}
\end{figure}

To capture holistic structures for possible high-resolution images, we should upsample the generated edges and lines to arbitrary scales without obvious degeneration. However, vanilla interpolation-based resizing causes zigzag as shown in Fig.~\ref{fig:upsample}(f)--(i). Such artifacts are more serious for large image sizes, which deteriorate the inpainted results. Fortunately, the grayscale sketch tensor is easy to be upsampled with a learning-based method. At first we train a simple CNN as the SSU to upsample edges and lines to a doubled size. Although lines can be upsampled successfully, edges fail to get correct results as shown in Fig.~\ref{fig:upsample}(j). Because there are ambiguities in the canny edge from different image sizes as shown in Fig.~\ref{fig:upsample}(b) and Fig.~\ref{fig:upsample}(c). 
Since lines got from a wireframe parser have good discrete representations~\cite{huang2018learning,xue2020holistically}, \emph{i.e.}, a line can be indicated as positions of two end-points and their relation, we can draw line maps in various resolutions without any ambiguities as shown in Fig.~\ref{fig:upsample}(d) and Fig.~\ref{fig:upsample}(e). If the model is trained in lines, it can also achieve smooth high-resolution edge maps due to the generalization of the network as shown in Fig.~\ref{fig:upsample}(k). Through iteratively calling the SSU, we can get high-quality edges and lines with high resolutions.

\subsection{ZeroRA Structure Enhanced Inpainting}
\label{sec:SFEandFTR}
\noindent\textbf{Fourier CNN Texture Restoration (FTR).} For the texture restoration, we adopt the excellent work of~\cite{suvorov2021resolution} as our inpainting backbone. Suvorov~\etal~\cite{suvorov2021resolution} propose to use Fourier convolutions~\cite{chi2020fast} for the frequency domain learning, which can achieve resolution-robust inpainted results. As backbones used by other inpainting models~\cite{nazeri2019edgeconnect,cao2021learning}, FTR is an autoencoder model with several convolutions for downsampling and upsampling image features. The key module of FTR is the Fast Fourier Convolution (FFC) layer, which is consisted of two branches: 1) the local branch uses conventional convolutions and 2) the global branch convolutes features after the fast Fourier transform. Then two branches are combined for larger receptive fields and local invariance during the inpainting~\cite{suvorov2021resolution}. However, such a powerful model fails to learn reasonable holistic structures. And we further propose a series of novel components to improve it.

\noindent\textbf{Structure Feature Encoder (SFE).}
For the given recovered edges $\tilde{\textbf{I}}_{e}$ and lines $\tilde{\textbf{I}}_{l}$ in arbitrary scales, we need a full convolutional network (FCN) to process them into a feature space. Our SFE is also an autoencoder model with 3 layers downsampling convolutions (encoder), 3 layers residual blocks with dilated convolutions~\cite{yu2015multi} (middle), and 3 layers upsampling convolutions (decoder). For the encoder and the decoder in SFE, we use gated convolutions (GCs)~\cite{yu2019free} to transfer useful features selectively. GC learns another sigmoid activation with the same channels. Then the sigmoid features are multiplied to the convoluted ones as outputs. Although GCs are widely used in image inpainting for the better generalization to irregular masks, we use GCs to filter useful features to FTR. Because the grayscale sketch space is sparse, and not all features are necessary for the inpainting. Then, 4 coarse-to-fine feature maps $\mathbf{S}_k, k\in\{0,1,2,3\}$ from the last middle layer and 3 decoder layers are selected to transfer structural features to FTR as
\vspace{-0.05in}
\begin{equation}
\vspace{-0.05in}
\mathbf{S}_0,\mathbf{S}_1,\mathbf{S}_2,\mathbf{S}_3=\mathrm{SFE}(\tilde{\textbf{I}}_{e},\tilde{\textbf{I}}_{l},\mathbf{M}),
\label{eq:sfe}
\end{equation}
where $\mathbf{M}$ indicates the resized binary mask.

\begin{figure}
\begin{centering}
\includegraphics[width=0.95\linewidth]{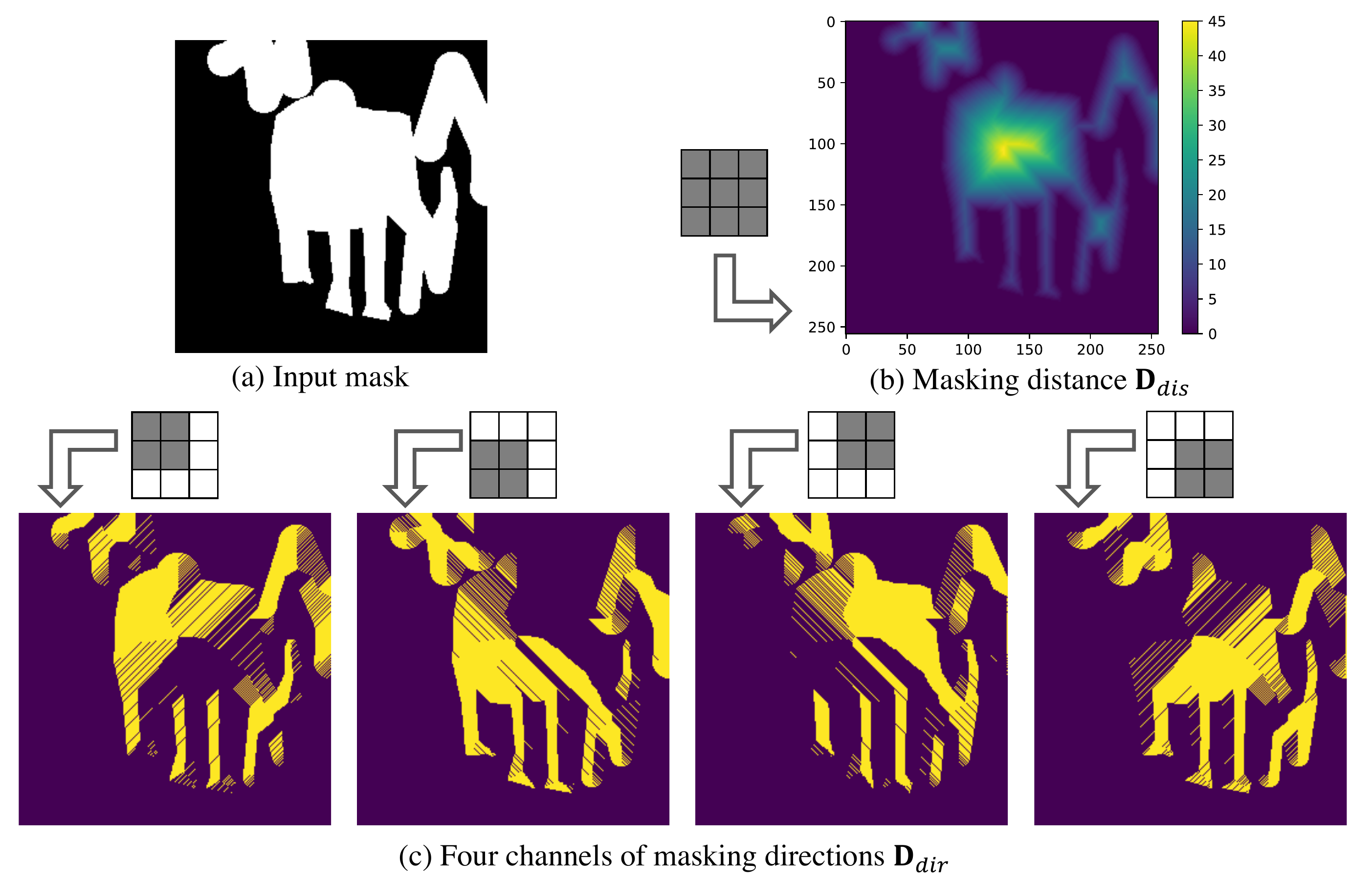}
\par\end{centering}
\vspace{-0.1in}
 \caption{The illustration of our masking relative position encoding. (a) Input mask, (b) masking distance $\mathbf{D}_{dis}$ and the all-one 3$\times$3 kernel, (c) masking directions $\mathbf{D}_{dir}$ and their kernels. \label{fig:MPE}}
 \vspace{-0.15in}

\end{figure}

\noindent\textbf{Masking Positional Encoding (MPE).}
Although the zero-padding in CNNs can provide some position information~\cite{islam2020much}, it only contains information about spatial anchors~\cite{xu2020positional}. Therefore, central generated regions from GANs tend to repeat meaningless artifacts without specific position encoding. When the image size is large, the effect of zero-padding will be further weakened, which causes more repeated artifacts~\cite{lin2021infinitygan} to generators.

During the inpainting, position information for unmasked regions is unnecessary, because the model always knows ground truth unmasked image regions. However, we think that position information is still critical for masked regions, especially when mask areas are large for high-resolution images. Limited by the receptive fields of CNNs, the model with large masks may lose the direction and position information, which causes meaningless artifacts. Although FFC can extend the feature learning to the frequency domain, it is insensitive to distinguish masked or unmasked regions. Therefore, we propose to use position encoding in masked regions called MPE for the image inpainting, which is orthogonal to and improves upon the FFC in FTR.

Specifically, to clearly represent masked and unmasked positional relations with our MPE written as $\mathbf{P}$, it can be expressed as the masking distance $\mathbf{P}_{dis}$ and the masking directions $\mathbf{P}_{dir}$ as shown in Fig.~\ref{fig:MPE}. 
Given an inversed 256$\times$256 binary mask, where one indicates unmasked regions and zero indicates masked regions, we use a 3$\times$3 all-one kernel to calculate the masking distance $\mathbf{D}_{dis}$ for each position in masked regions as shown in Fig.~\ref{fig:MPE}(b). Then, the distance is clipped and mapped by the Sinusoidal Positional Encoding (SPE)~\cite{vaswani2017attention} to get $\mathbf{P}_{dis}\in\mathbb{R}^{256\times256\times d}$
\vspace{-0.05in}
\begin{equation}
\vspace{-0.05in}
\begin{split}
\mathbf{P}_{dis,2i}&=\mathrm{sin}(\mathrm{clip}(\mathbf{D}_{dis},0,D_{max})/10000^{\frac{i}{d}}),\\
\mathbf{P}_{dis,2i+1}&=\mathrm{cos}(\mathrm{clip}(\mathbf{D}_{dis},0,D_{max})/10000^{\frac{i}{d}}),
\end{split}
\label{eq:masking_distance}
\end{equation}
where $i$ indicates the channel index; $D_{max}=128$, and $d=64$ means the total channels of $\mathbf{P}_{dis}$, which is the same as the first convolution of FTR. Since SPE can only provide absolute positional information~\cite{xu2020positional}, $\mathbf{P}_{dis}$ can be further resized by the nearest interpolation to various scales during the training for learning relative positional information in arbitrary resolutions.
For masking directions, we use 4 different binary kernels to get the 4-channel one-hot vector $\mathbf{D}_{dir}\in\mathbb{R}^{256\times256\times4}$. Values of $\mathbf{D}_{dir}$ depend on which kernel covers the masked regions firstly. $\mathbf{D}_{dir}$ shows the nearest direction from a masked position to an unmasked one as shown in Fig.~\ref{fig:MPE}(c). 
Note that the masking direction is a multi-label vector, because a pixel may have more than one shortest direction. Then $\mathbf{D}_{dir}$ is projected to a $d$ dimension features with learnable embedding parameters $\mathbf{W}_{dir}\in\mathbb{R}^{4\times d}$ as
\vspace{-0.05in}
\begin{equation}
\vspace{-0.05in}
\mathbf{P}_{dir}=\mathbf{D}_{dir}\times\mathbf{W}_{dir}\in\mathbb{R}^{256\times256\times d}.
\label{eq:masking_direction}
\end{equation}
$\mathbf{P}_{dis}$ and $\mathbf{P}_{dir}$ are added as MPE to the first layer of FTR.

\noindent\textbf{Zero-initialized Residual Addition (ZeroRA).}
Since most inpainting methods are based on sophisticated GANs nowadays, training the inpainting model incrementally is non-trivial. However, benefiting from various auxiliary information~\cite{nazeri2019edgeconnect,cao2021learning,liao2020guidance}, incrementally training is flexible to improve the image inpainting. 
To improve the pretrained inpainting model incrementally with holistic structures, we propose to use ZeroRA, which has been leveraged in~\cite{bachlechner2020rezero} to replace the layer normalization in the transformer. The idea of ZeroRA is simple. For the given input feature $x$, the output feature $x'$ is got from adding a skip connection with function $F$ to $x$ with a zero-initialized trainable residual weight $\alpha$ as
\vspace{-0.05in}
\begin{equation}
\vspace{-0.05in}
x'=x+\alpha\cdot F(x).
\label{eq:rezero}
\end{equation}
For simple linear-based models, if $\alpha$ is initialized in zero, the input-output Jacobian will be initialized to 1, which makes the training stable. For more complex cases, experiments in~\cite{bachlechner2020rezero} also prove the effectiveness of ZeroRA. Since ZeroRA can replace the layer normalization in the transformer, it can also improve the expressive power of the model without degrading variances to early layers.

In our case, we use ZeroRA to incrementally add structural information from SFE to FTR. Specifically, 4 zero-initialized $\alpha_k, k\in\{0,1,2,3\}$ are utilized to fuse 4 related feature maps $\mathbf{S}_k$ from SFE. For the feature $\mathbf{X}_k$ of FTR encoder layer $k$, which is based on Conv-BatchNorm-ReLU, we add residuals as follows
\vspace{-0.05in}
\begin{equation}
\vspace{-0.05in}
\begin{split}
\mathbf{X}_{k+1}&=\mathrm{Conv}(\mathbf{X}_k+\alpha_k\cdot\mathbf{S}_k),\\
\mathbf{X}_{k+1}&=\mathrm{ReLU}(\mathrm{BatchNorm}(\mathbf{X}_{k+1})).
\end{split}
\label{eq:rezero_ftr}
\end{equation}
There is another advantage of the ZeroRA based incremental learning. The model output is equivalent to the pretrained one at the beginning of finetuning, which can effectively stable the training, and transfer necessary information adaptively. Our ablation studies show that the ZeroRA is important to incrementally finetune the pretrained inpainting model with additional information.

\subsection{Loss Functions}
We adopt the same loss functions as~\cite{suvorov2021resolution}, which include L1 loss, adversarial loss, feature match loss, and high receptive field  (HRF) perceptual loss~\cite{suvorov2021resolution}.
Firstly, L1 loss is only calculated between the unmasked regions as
\vspace{-0.05in}
\begin{equation}
\vspace{-0.05in}
\mathcal{L}_{L1}=(1-\mathbf{M})\odot|\mathbf{\hat{I}}-\mathbf{\tilde{I}}|_1,
\label{eq:l1_loss}
\end{equation}
where $\mathbf{M}$ indicates 0-1 mask that 1 means masked regions; $\odot$ means the element-wise multiplication; $\mathbf{\hat{I}},\mathbf{\tilde{I}}$ indicate the ground truth and predicted images respectively. The adversarial loss is consisted of the discriminator loss $\mathcal{L}_D$ and the generator loss $\mathcal{L}_G$. Moreover, we only regard features from masked regions as fake samples in $\mathcal{L}_D$. The PatchGAN~\cite{isola2017image} based discriminator is written as $D$ and the combination of FTR and SFE can be seen as the generator $G$, Then the adversarial loss can be indicated as
\vspace{-0.05in}
\begin{equation}
\vspace{-0.05in}
\begin{split}
\negthickspace\negthickspace
\mathcal{L}_{D}=&-\mathbb{E}_{\mathbf{\hat{I}}}\left[\mathrm{log}D(\mathbf{\hat{I}})\right]-\mathbb{E}_{\mathbf{\tilde{I}},\mathbf{M}}\left[\mathrm{log}D(\mathbf{\tilde{I}})\odot(1-\mathbf{M})\right]\\
&-\mathbb{E}_{\mathbf{\tilde{I}},\mathbf{M}}\left[\mathrm{log}(1-D(\mathbf{\tilde{I}}))\odot\mathbf{M}\right],\\
&\mathcal{L}_{G}=-\mathbb{E}_{\mathbf{\tilde{I}}}\left[\mathrm{log}D(\mathbf{\tilde{I}})\right],\\
&\mathcal{L}_{adv}=\mathcal{L}_D+\mathcal{L}_G+\lambda_{GP}\mathcal{L}_{GP},
\end{split}
\label{eq:adv_loss}
\end{equation}
where $\mathcal{L}_{GP}=\mathbb{E}_{\mathbf{\hat{I}}}||\triangledown_{\mathbf{\hat{I}}}D(\mathbf{\hat{I}})||^2$ is the gradient penalty~\cite{gulrajani2017improved} and $\lambda_{GP}=1e-3$. We also use the feature match loss~\cite{wang2018high} $\mathcal{L}_{fm}$, which is based on L1 loss between discriminator features of true and fake samples. $\mathcal{L}_{fm}$ is usually used to stable the GAN training. It can also slightly improve the performance. 
Furthermore, we use the HRF loss $\mathcal{L}_{hrf}$ in~\cite{suvorov2021resolution} as
\vspace{-0.05in}
\begin{equation}
\vspace{-0.05in}
\mathcal{L}_{hrf}=\mathbb{E}(\left[\phi_{hrf}(\mathbf{\hat{I}})-\phi_{hrf}(\mathbf{\tilde{I}})\right]^2),
\label{eq:hrfpl}
\end{equation}
where $\phi_{hrf}$ indicates a pretrained segmentation ResNet50 with dilated convolutions.
As discussed in~\cite{suvorov2021resolution}, using HRF loss instead of the perceptual loss~\cite{johnson2016perceptual} can improve the quality of the inpainting model. The final loss of our model in the incremental training can be written as
\vspace{-0.05in}
\begin{equation}
\vspace{-0.05in}
\small
\begin{split}
\negthickspace
\mathcal{L}_{final}=&\lambda_{L1}\mathcal{L}_{L1}+\lambda_{adv}\mathcal{L}_{adv}+\lambda_{fm}\mathcal{L}_{fm}+\lambda_{hrf}\mathcal{L}_{hrf},
\end{split}
\label{eq:final_loss}
\end{equation}
where $\lambda_{L1}=10,\lambda_{adv}=10,\lambda_{fm}=100,\lambda_{hrf}=30$.

\section{Experiments}

\subsection{Datasets}

The proposed ZITS is trained on two datasets: Places2~\cite{zhou2017places} and our custom indoor dataset (Indoor). For Places2, we use about 1,800k images from various scenes as the training set, and 36,500 images as the validation. To better demonstrate the structural recovery, we collect 5,000 images from ShanghaiTech~\cite{huang2018learning} and 15,055 images from NYUDepthV2~\cite{Silberman:ECCV12} to build the custom 20,055 Indoor training dataset. For the Indoor validation, we collect 1,000 images which are consist of 462 and 538 images from ShanhaiTech and NYUDepthV2 respectively. Places2 and Indoor can all be tested in both 256$\times$256 and 512$\times$512. Besides, we also test the inpainting ability on high-resolution MatterPort3D~\cite{chang2017matterport3d} with 1,965 indoor images in 1024$\times$1024. More details and results of MatterPort3D are discussed in the supplementary.


\subsection{Implementation Details}

\noindent\textbf{Training Settings.}
Our ZITS is implemented with PyTorch. For the training of TSR, we use the Adam optimizer of learning rate 6e-4 with 1,000 steps warmup and cosine decay. TSR is trained with 150k and 400k steps for Indoor and Places2. On the other hand, we first train the FTR with Adam optimizer of learning rates 1e-3 and 1e-4 for generator and discriminator respectively. And FTR is trained with 100k steps on Indoor and 800k steps on Places2. Then, we incrementally finetune them with ZeroAR for just 50k steps on both Indoor and Places2, and reduce the generator learning rate to 3e-4. Besides, we warmup the learning rate for training the SFE with 2,000 steps.
For the training of TSR and FTR, input images are resized into 256$\times$256. For the incremental finetuning, we separately train two versions of ZITS, which are the version trained in 256$\times$256 and the version trained in random size from 256 to 512. The second model can handle some situations with higher resolution inputs. And the MPE is also changed to a relative position encoding for the random size training.

\noindent\textbf{Mask Settings.}
To tackle the real-world object removal task, we follow the mask setting from~\cite{cao2021learning}, which includes irregular masking brushes and COCO~\cite{lin2014microsoft} segmentation masks with masking rates from 10\% to 50\%. Different from~\cite{cao2021learning}, we randomly combine irregular and segmentation masks with 20\% to improve the learning difficulty. 

\subsection{Comparison Methods}
We compare the proposed model with other state-of-the-art methods, which include Edge Connect (EC)~\cite{nazeri2019edgeconnect}, Contextual Residual Aggregation (HiFill)~\cite{yi2020contextual}, Multi-scale Sketch Tensor inpainting (MST)~\cite{cao2021learning}, Co-Modulation GAN (Co-Mod)~\cite{zhao2021large}, and Large Mask inpainting (LaMa)~\cite{suvorov2021resolution}. All competitors are compared in the Places2. We also retrain EC, MST, and LaMa for the Indoor dataset to discuss the structure recovery. Note that the LaMa compared below are all trained with the same total steps as ZITS.


\subsection{Quantitative Comparisons}

\begin{table}
\caption{Quantitative results on Indoor and Places2 in 256$\times$256. \label{table:main_results}}
\vspace{-0.15in}
\begin{centering}
\renewcommand\tabcolsep{2.0pt}
\footnotesize
\begin{tabular}{c|cccc|cccc}
\hline 
\multicolumn{5}{c|}{Indoor} & \multicolumn{4}{c}{Places2}\tabularnewline
\hline 
\hline 
 & PSNR$\uparrow$ & SSIM$\uparrow$ & FID$\text{\ensuremath{\downarrow}}$ & LPIPS$\text{\ensuremath{\downarrow}}$ & PSNR$\uparrow$ & SSIM$\uparrow$ & FID$\text{\ensuremath{\downarrow}}$ & LPIPS$\text{\ensuremath{\downarrow}}$\tabularnewline
\hline 
EC & 24.07 & 0.884 & 22.02 & 0.135 & 23.31 & 0.839 & 6.21 & 0.149\tabularnewline
\hline 
MST & 24.52 & 0.894 & 21.65 & 0.122 & 24.02 & 0.862 & 3.53 & 0.137\tabularnewline
\hline 
HiFill & - & - & - & - & 20.76 & 0.770 & 21.33 & 0.246\tabularnewline
\hline 
Co-Mod & - & - & - & - & 22.57 & 0.843 & 1.49 & 0.122\tabularnewline
\hline 
LaMa & 25.20 & 0.902 & 16.97 & 0.112 & 24.37 & 0.869 & 1.63 & 0.155\tabularnewline
\hline 
Ours & \textbf{25.57} & \textbf{0.907} & \textbf{15.93} & \textbf{0.098} & \textbf{24.42} & \textbf{0.870} & \textbf{1.47} & \textbf{0.108}\tabularnewline
\hline 
\end{tabular}
\par\end{centering}
\vspace{-0.15in}
\end{table}

\noindent\textbf{Inpainting Results.} 
In Tab.~\ref{table:main_results}, we utilize PSNR, SSIM~\cite{wang2004image}, FID~\cite{heusel2018gans}, and LPIPS~\cite{zhang2018unreasonable} to assess the performance of all compared methods on the Indoor and Places2 datasets in 256$\times$256 with mixed segmentation and irregular masks. 
More results with different masking rates are shown in the supplementary.
For Indoor, our ZITS can achieve the best results on all metrics. While MST can get slightly better results compared with EC, which is benefited by the usage of lines. 
LaMa can get more acceptable FID and LPIPS while our ZITS can achieve significant improvements based on LaMa due to the seamlessly embedded structural information and positional encoding.
Note that the gap between ZITS and MST is also caused by the quality gap of the structure recovery as discussed below. 
For Places2, HiFill fails to get good results with large masks, which may be caused by its limited model capacity. Note that Co-Mod has a low FID and LPIPS on Places2. However, Co-Mod is trained with a sophisticated StyleGAN~\cite{karras2020analyzing} with much more training data compared with others. And our ZITS can even achieve slightly better results compared with Co-Mod with limited data scale and training steps. In general, our method has superior performance compared with LaMa, which is valuable with only 50k finetune steps. And LaMa in Tab.~\ref{table:main_results} are trained with the same total steps as ZITS.

\begin{table}
\caption{Quantitative Precision (P.), Recall (R.) and F1-score (F1) of edges and lines on Indoor and Places2. \label{table:edgeline_res}} 
 \vspace{-0.15in}
\begin{centering}
\renewcommand\tabcolsep{2.8pt}
\small
\begin{tabular}{c|c|ccc|ccc|c}
\hline 
\multicolumn{1}{c}{} &  & \multicolumn{3}{c|}{Edge} & \multicolumn{3}{c|}{Line} & Avg\tabularnewline
\multicolumn{1}{c}{} &  & P. & R. & F1 & P. & R. & F1 & F1\tabularnewline
\hline 
\hline 
\multirow{2}{*}{Indoor} & MST & 23.79 & 26.87 & 21.36 & 43.67 & 51.95 & 37.77 & 33.73\tabularnewline
 & Ours & \textbf{37.34} & \textbf{34.25} & \textbf{35.10} & \textbf{53.60} & \textbf{66.23} & \textbf{58.35} & \textbf{46.72}\tabularnewline
\hline 
\multirow{2}{*}{Places2} & MST & 22.54 & 18.29 & 20.19 & 34.22 & 49.21 & 37.09 & 28.64\tabularnewline
 & Ours & \textbf{35.64} & \textbf{27.92} & \textbf{30.39} & \textbf{43.70} & \textbf{60.54} & \textbf{49.35} & \textbf{39.87}\tabularnewline
\hline 
\end{tabular}
\par\end{centering}
 \vspace{-0.1in}
\end{table}

\noindent\textbf{Results of Edges and Lines.} 
We show quantitative results of edges and lines on Indoor and Places2 in Tab.~\ref{table:edgeline_res}. Our TSR can achieve much better results on both Indoor and Places2 compared with MST. It demonstrates that the transformer-based TSR is amenable to learning holistic structures in a sparse tensor space, which can benefit the results of ZITS a lot as shown in Tab.~\ref{table:main_results}. Note that TSR results in Tab.~\ref{table:edgeline_res} are based on Mask-Predict~\cite{cho2020xlxmert, guo2020incorporating, ghazvininejad2019mask}, which can enrich the structural generation by iteratively sampling outputs but does not improve the quantitative metrics. More about Mask-Predict are discussed in the supplementary.

\subsection{Qualitative Comparisons}

\begin{figure*}
\begin{centering}
\includegraphics[width=0.8\linewidth]{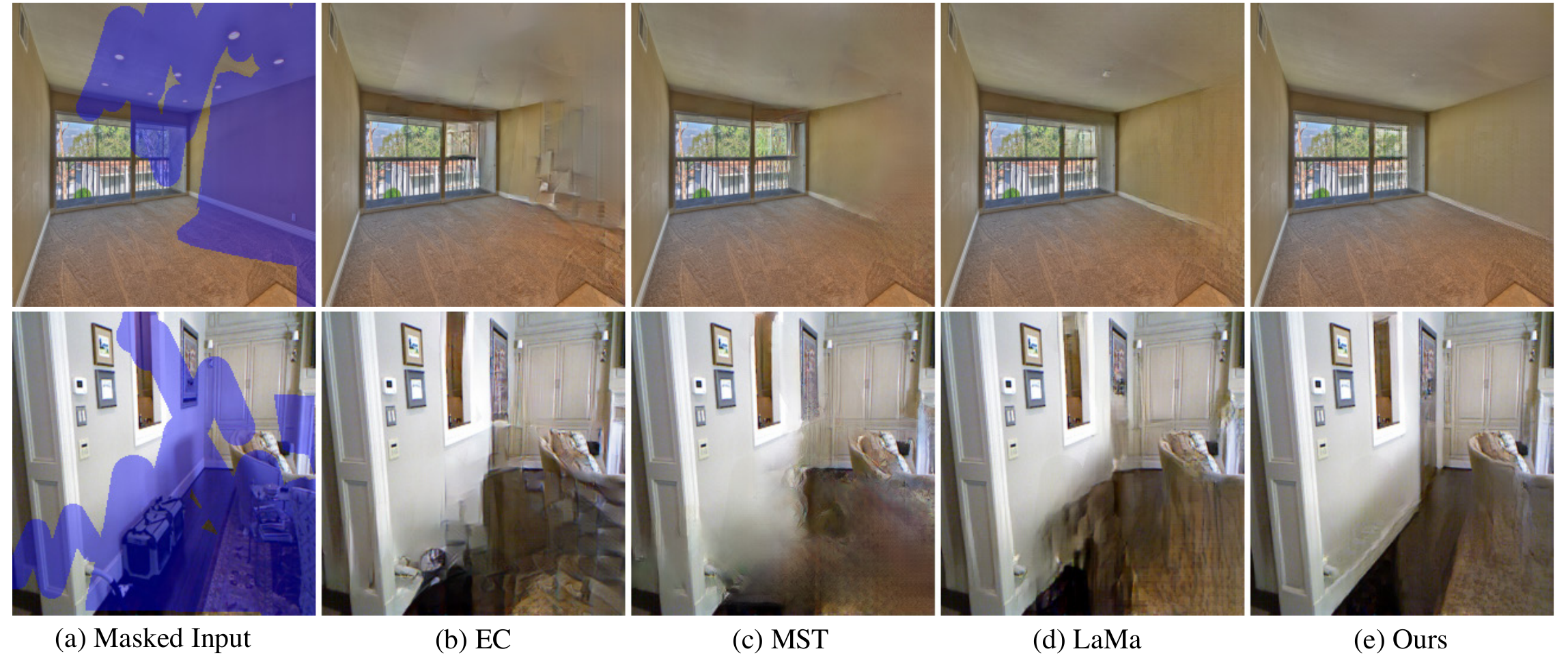}
\par\end{centering}
 \vspace{-0.15in}
 \caption{Qualitative results of Indoor dataset compared among EC~\cite{nazeri2019edgeconnect}, MST~\cite{cao2021learning}, LaMa~\cite{suvorov2021resolution}, and ours. Zoom-in for details. \label{fig:qualitative_indoor}}
 \vspace{-0.1in}
\end{figure*}

\begin{figure*}
\begin{centering}
\includegraphics[width=0.85\linewidth]{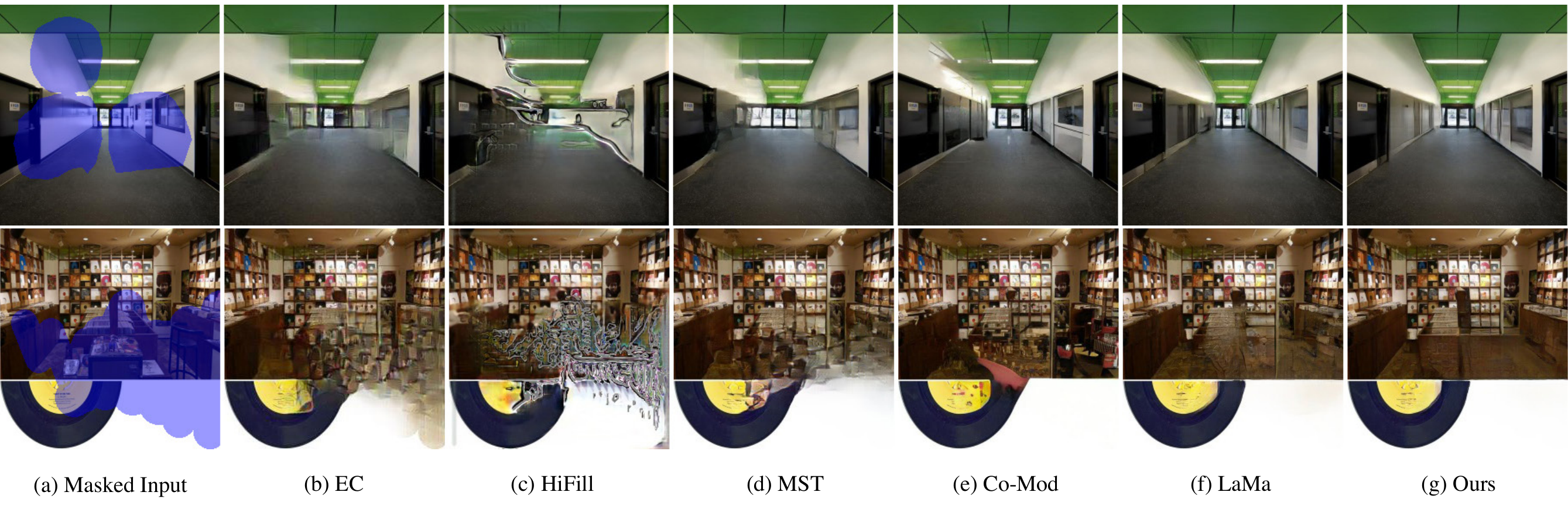}
\par\end{centering}
 \vspace{-0.15in}
 \caption{Qualitative results of Places2 compared among EC~\cite{nazeri2019edgeconnect}, HiFill~\cite{yi2020contextual}, MST~\cite{cao2021learning}, Co-Mod~\cite{zhao2021large}, LaMa~\cite{suvorov2021resolution}, and ours. \label{fig:qualitative_places2}}
 \vspace{-0.1in}

\end{figure*}

\noindent\textbf{Inpainting Results.} We show the qualitative inpainting results of Indoor in Fig.~\ref{fig:qualitative_indoor} and Places2 in Fig.~\ref{fig:qualitative_places2}. Compared with other methods, our ZITS can tackle more reasonable structures, especially our method can obtain clearer borderlines. Furthermore, ZITS achieves prominent improvements in the structure recovery compared with LaMa. Note that both LaMa and ZITS are trained with the same steps.

\begin{figure}
\begin{centering}
\includegraphics[width=0.95\linewidth]{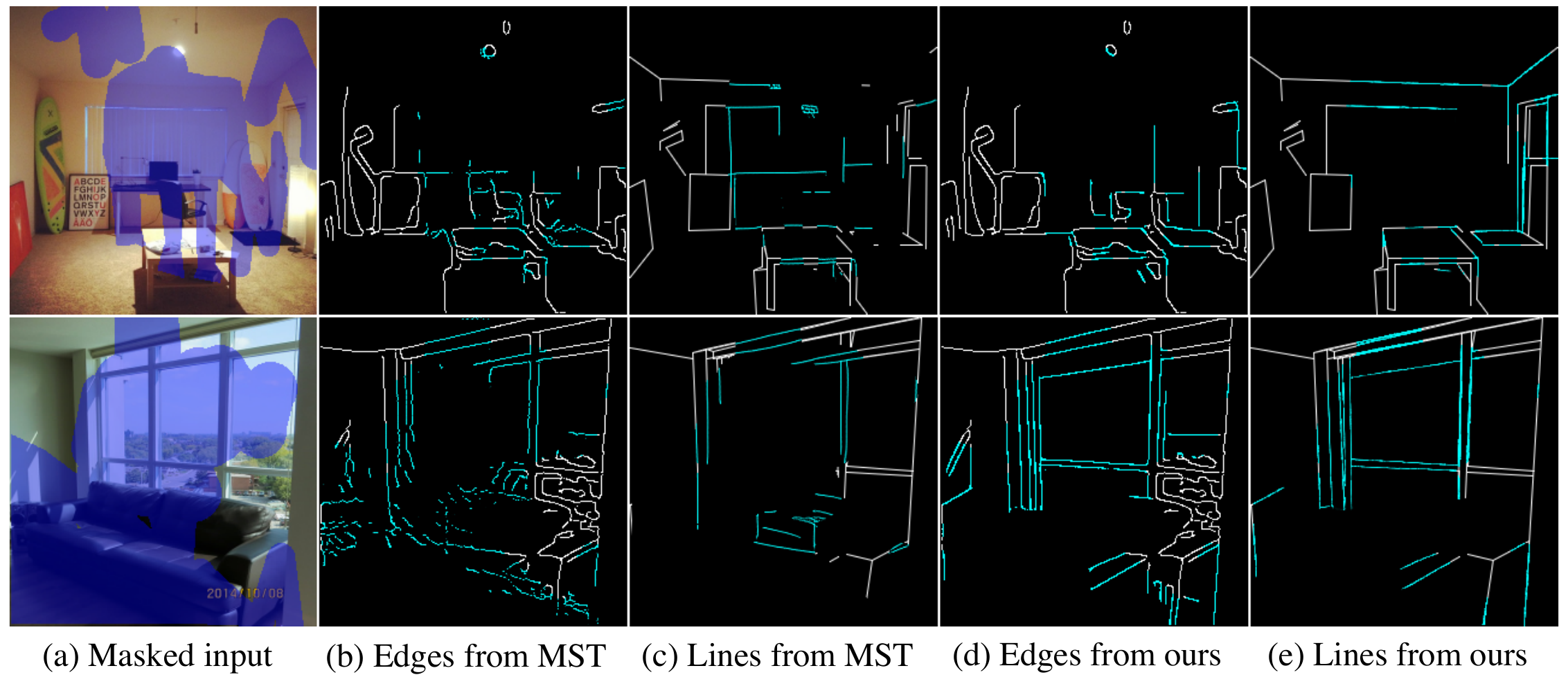}
\par\end{centering}
 \vspace{-0.15in}
 \caption{Qualitative results of edges and lines of Indoor dataset.
 \label{fig:qualitative_str}}
 \vspace{-0.15in}
\end{figure}

\noindent\textbf{Results of Edges and Lines.} We compared the structure recovery results in Indoor of Fig.~\ref{fig:qualitative_str}, which are compared between our transformer-based TSR and CNN-based model from MST. Our TSR can achieve more reasonable and expressive results of both edges and lines. More qualitative structural results are shown in the supplementary.

\begin{figure}
\begin{centering}
\includegraphics[width=1.0\linewidth]{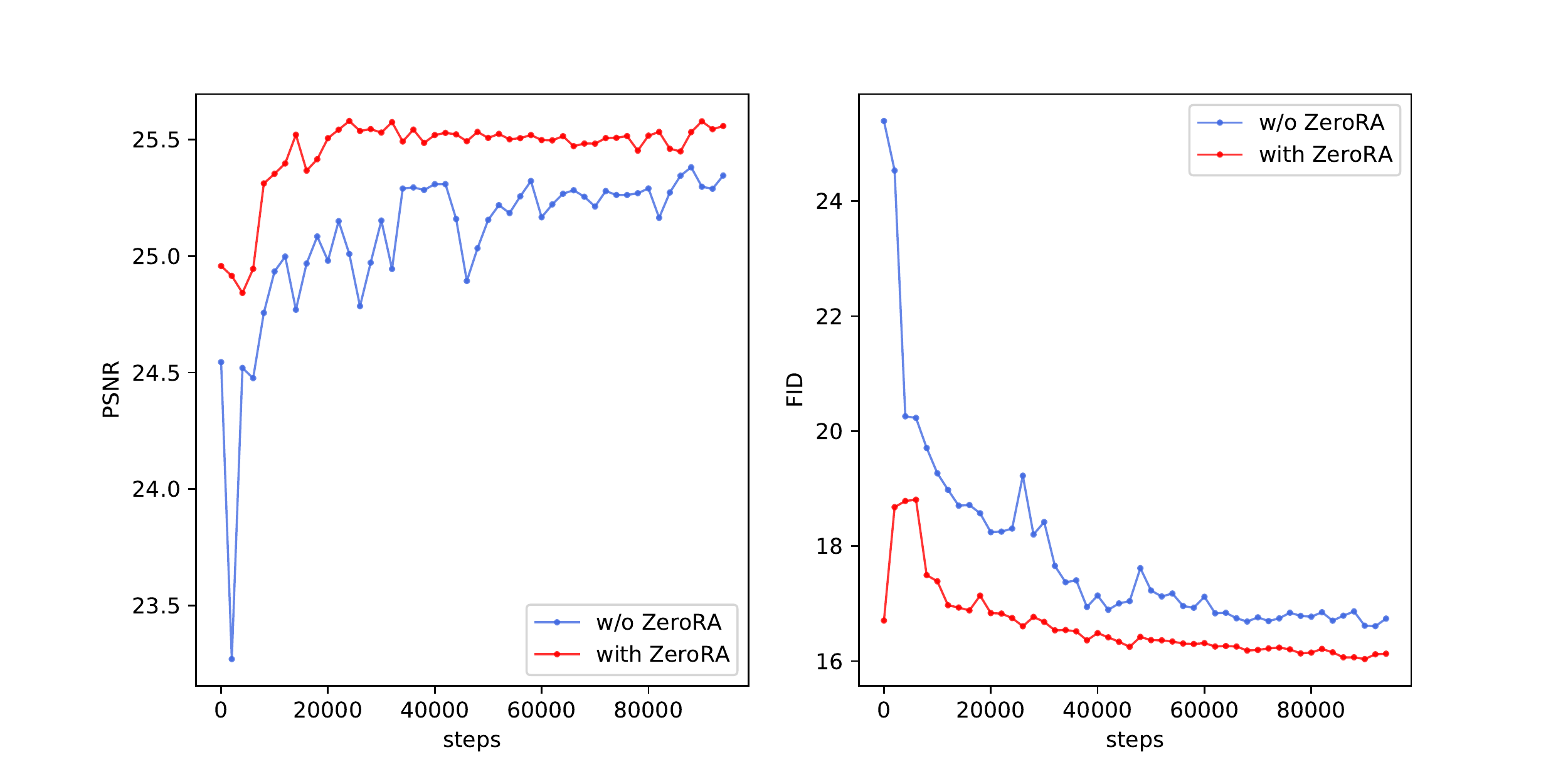}
\par\end{centering}
 \vspace{-0.15in}
 \caption{Line charts of the structural enhanced finetuning with and without ZeroRA.
 \label{fig:zerora_abla}}
 \vspace{-0.15in}
\end{figure}

\begin{figure}
\begin{centering}
\includegraphics[width=0.75\linewidth]{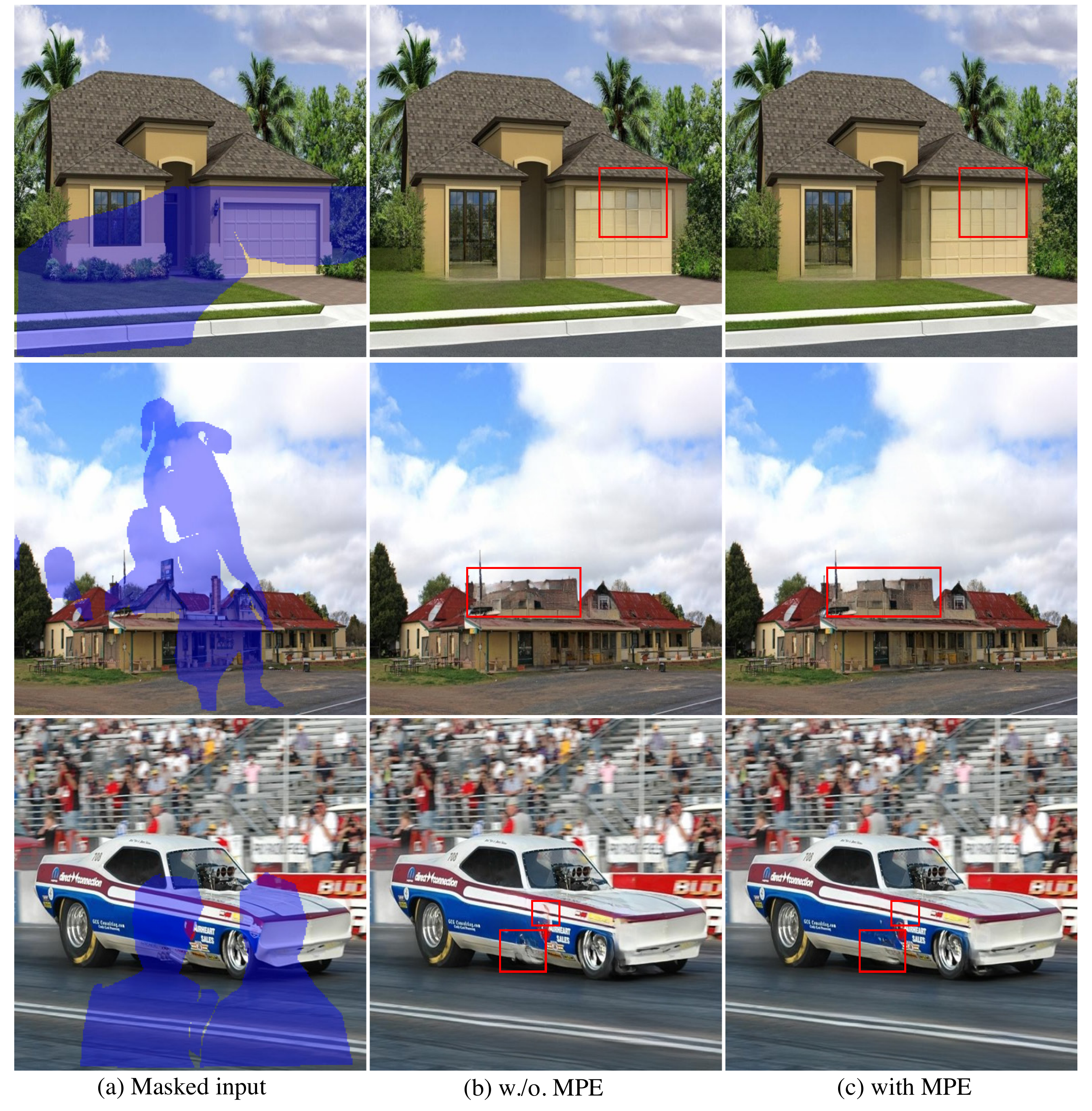}
\par\end{centering}
 \vspace{-0.15in}
 \caption{Ablations of 512$\times$512 Places2 with and without MPE.
 \label{fig:MPE_abla}}
 \vspace{-0.15in}
\end{figure}

\begin{table}
\small
\caption{Ablation studies of MPE on 512$\times$512 Places2 finetuned with dynamic resolutions from 256 to 512.\label{table:abla_MPE}}
 \vspace{-0.15in}
\centering
\begin{tabular}{c|cccc}
\hline 
 & PSNR$\uparrow$ & SSIM$\uparrow$ & FID$\downarrow$ & LPIPS$\downarrow$\tabularnewline
\hline 
\hline 
with MPE & \textbf{24.23} & \textbf{0.881} & \textbf{26.08} & \textbf{0.133}\tabularnewline
w./o. MPE & 24.20 & 0.880 & 26.29 & 0.135 \tabularnewline
\hline 
\end{tabular}
\vspace{-0.15in}
\end{table}

\begin{table}
\small
\caption{Ablation studies with different settings on Indoor. \label{table:ablation}}
 \vspace{-0.15in}
\renewcommand\tabcolsep{1.5pt}
\begin{centering}
\begin{tabular}{ccccc|cccc}
\hline 
FTR & SFE & MPE & ReZero & GateConv & PSNR$\uparrow$ & SSIM$\uparrow$ & FID$\downarrow$ & LPIPS$\downarrow$\tabularnewline
\hline 
\hline 
\CheckmarkBold{} &  &  &  &  & 25.20 & 0.902 & 16.97 & 0.112\tabularnewline
\CheckmarkBold{} &  & \CheckmarkBold{} &  &  & 25.31 & 0.903 & 16.44 & 0.110\tabularnewline
\CheckmarkBold{} & \CheckmarkBold{} & \CheckmarkBold{} &  & \CheckmarkBold{} & 25.28 & 0.905 & 16.15 & 0.102\tabularnewline
\CheckmarkBold{} & \CheckmarkBold{} &  & \CheckmarkBold{} & \CheckmarkBold{} & 25.46 & 0.906 & 16.22 & 0.107\tabularnewline
\CheckmarkBold{} & \CheckmarkBold{} & \CheckmarkBold{} & \CheckmarkBold{} &  & 25.51 & 0.906 & 16.15 & 0.103\tabularnewline
\CheckmarkBold{} & \CheckmarkBold{} & \CheckmarkBold{} & \CheckmarkBold{} & \CheckmarkBold{} & \textbf{25.57} & \textbf{0.907} & \textbf{15.93} & \textbf{0.098}\tabularnewline
\hline 
\end{tabular}
\par\end{centering}
 \vspace{-0.2in}
\end{table}

\subsection{Ablation Studies}
\label{sec:ablation}
Quantitative ablation studies on Indoor are shown in Tab.~\ref{table:ablation}. MPE and GCs can slightly improve the performance of FTR. Besides, if adding structural information from TSR without ZeroRA, the improvement is limited. So ZeroRA is useful for incremental learning with a good convergence. Moreover, the full model achieves the best performance.

\noindent\textbf{ZeroRA.} We also show line charts of PSNR and FID during the finetuning in Fig.~\ref{fig:zerora_abla} with and without ZeroRA. The blue curve without ZeroRA is unstable at the beginning of the finetuning, while the red one with ZeroRA enjoys better convergence and stability. Because adding extra structural features without ZeroRA leads to dramatic output changing, which harms the vulnerable GAN training.

\noindent\textbf{MPE.} We further exploit the effects of MPE in high-resolution inpainting. 
FTR is trained without MPE at first. Then we use the ZeroRA technique to finetune the model with and without MPE of the same steps. Results in Tab.~\ref{table:abla_MPE} show that the simple MPE based finetuning effectively improves the 512-inpainting in FID. From Fig.~\ref{fig:MPE_abla}, ZITS with MPE generates images with natural and smooth colors.




\subsection{Results of High-Resolution Inpainting}

We also compare the results of HiFill, Co-Mod, LaMa, and our ZITS in Places2(512) in Tab.~\ref{table:high_res}. Besides, LaMa and ZITS are further compared in Indoor(512) and MatterPort3D(1k) in Tab.~\ref{table:high_res}. LaMa and ZITS are firstly trained in 256$\times$256 and then finetuned with dynamic resolutions from 256 to 512 with 50k steps. Models tested in Indoor(512) and MatterPort3D(1k) are both trained in Indoor training set. For the Places2(512), we randomly select 1,000 samples from 36,500 for the 512 testing. Our ZITS can achieve prominent improvements compared with LaMa, which illustrates that our MPE and incremental structure enhanced training is effective for high-resolution inpainting. Besides, ZITS can also get better 1k results in MatterPort3D. More high-resolution results can be seen in the supplementary.

\begin{table}
\small
\vspace{+0.05in}
\caption{Quantitative results of 512$\times$512 in Indoor and Places2, and 1024$\times$1024 MatterPort3D. \label{table:high_res}}
 \vspace{-0.15in}
\renewcommand\tabcolsep{2.8pt}
\begin{centering}
\begin{tabular}{c|c|cccc}
\hline 
\multicolumn{1}{c}{} &  & PSNR$\uparrow$ & SSIM$\uparrow$ & FID$\downarrow$ & LPIPS$\downarrow$\tabularnewline
\hline 
\hline 
\multirow{2}{*}{Indoor(512)} & LaMa & 24.42 & 0.911 & 21.48 & 0.826 \tabularnewline
 & Ours & \textbf{25.36} & \textbf{0.919} & \textbf{18.76} & \textbf{0.823} \tabularnewline
\hline 
\multirow{4}{*}{Places2(512)} & HiFill & 20.10 & 0.764 & 65.47 & 0.291\tabularnewline
 & Co-Mod & 22.00 & 0.843 & 30.04 & 0.166\tabularnewline
 & LaMa & 24.15 & 0.877 & 27.86 & 0.149\tabularnewline
 & Ours & \textbf{24.23} & \textbf{0.881} & \textbf{26.08} & \textbf{0.133}\tabularnewline
\hline 
\multirow{2}{*}{MatterPort3D(1k)} & LaMa & 26.40 & 0.944 & 14.04 & 0.133 \tabularnewline
 & Ours & \textbf{26.55} & \textbf{0.946} & \textbf{12.34} & \textbf{0.116} \tabularnewline
\hline 
\end{tabular}
\par\end{centering}
 \vspace{-0.15in}
\end{table}

\section{Conclusions}

In this paper, we propose an incrementally structure enhanced inpainting model called ZITS. We use a transformer-based structure restorer to get much better holistic structures compared with previous methods. Then, a novel ZeroRA strategy is leveraged to incorporate auxiliary structures into a pretrained inpainting model with a few finetuning steps. The proposed masking positional encoding can further improve the inpainting performance. ZITS can achieve significant improvements based on the state-of-the-art model in experiments of various resolutions.

\appendix

\section{Broader Impacts}
All generated results of both the main paper and the supplementary are based on learned statistics of the training dataset. Therefore, the results only reflect biases in those data without our subjective opinion. This work is only researched for the algorithmic discussion, and related societal impacts should not be ignored by users. 

\section{Detailed Network Settings}

We show some detailed network settings in Tab.~\ref{table:network_settings}. Besides, the transformer block and Fast Fourier Convolution (FFC) block~\cite{suvorov2021resolution} have been introduced in the main paper. The dilated resnet block is from the middle layer of~\cite{nazeri2019edgeconnect} with dilate=2. 

\begin{table*}
\caption{Model settings of Transformer Structure Restoration (TSR), Structure Feature Encoder (SFE), and Fourier CNN Texture Restoration (FTR). GC, BN mean Gated Convolution~\cite{yu2019free} and BatchNorm; TConv2d, TGC indicate Transposed Conv2d and GC.
\label{table:network_settings}}
\begin{centering}
\small
\begin{tabular}{c|c|c}
\hline 
Transformer Structure Restoration (TSR) & Structure Feature Encoder (SFE) & Fourier CNN Texture Restoration (FTR)\tabularnewline
\hline 
\hline 
Conv2d+ReLU($256\times256\times64$) & GC+BN+ReLU($256\times256\times64$) & Conv2d+BN+ReLU($256\times256\times64$)\tabularnewline
\hline 
Conv2d+ReLU($128\times128\times128$) & GC+BN+ReLU($128\times128\times128$) & Conv2d+BN+ReLU($128\times128\times128$)\tabularnewline
\hline 
Conv2d+ReLU($64\times64\times256$) & GC+BN+ReLU($64\times64\times256$) & Conv2d+BN+ReLU($64\times64\times256$)\tabularnewline
\hline 
Conv2d+ReLU($32\times32\times256$) & GC+BN+ReLU($32\times32\times512$) & Conv2d+BN+ReLU($32\times32\times512$)\tabularnewline
\hline 
TransformerBlock$\times8$ & DilatedResnetBlock$\times3$ & FFCBlock$\times9$\tabularnewline
\hline 
TConv2d+ReLU($64\times64\times256$) & TGC+BN+ReLU($64\times64\times256$) & TConv2d+BN+ReLU($64\times64\times256$)\tabularnewline
\hline 
TConv2d+ReLU($128\times128\times128$) & TGC+BN+ReLU($128\times128\times128$) & TConv2d+BN+ReLU($128\times128\times128$)\tabularnewline
\hline 
TConv2d+ReLU($256\times256\times64$) & TGC+BN+ReLU($256\times256\times64$) & TConv2d+BN+ReLU($256\times256\times64$)\tabularnewline
\hline 
Conv2d+Sigmoid($256\times256\times2$) & -- & Conv2d+Tanh($256\times256\times3$)\tabularnewline
\hline 
\end{tabular}
\par\end{centering}
\end{table*}

\section{More Training Details}

Training a model with dynamic resolutions of 256$\sim$512 reduces the training speed with frequent GPU memory swaps. Therefore, we train the model with regular resolutions, \emph{i.e.}, resizing images from 512 to 256 and then back to 512. For Indoor, there is one cycle for each epoch. For Places2, there are 64 cycles for each epoch. Such a local monotonic resizing makes the training smooth without missing diversity. And the dynamic resolution based training can effectively save the training cost compared with the training with a full 512 image size. Moreover, it benefits to learn relative position encoding for our proposed MPE as discussed in~\cite{xu2020positional}.

Our TSR can be trained in batch size 30 with 3 NVIDIA(R) Tesla(R) V100 16GB GPUs. 256$\times$256 based FTR and SFE can be trained in batch size 30 with 3 V100 GPUs. For the dynamic resolution based training, we use batch size 18 with 6 V100 GPUs. The ZeroRA based finetuning cost only about half a day and one day for 256$\times$256 and 256$\sim$512 resolutions respectively.

\section{Upsampling Iteratively with SSU}

\begin{figure*}
\begin{centering}
\includegraphics[width=0.99\linewidth]{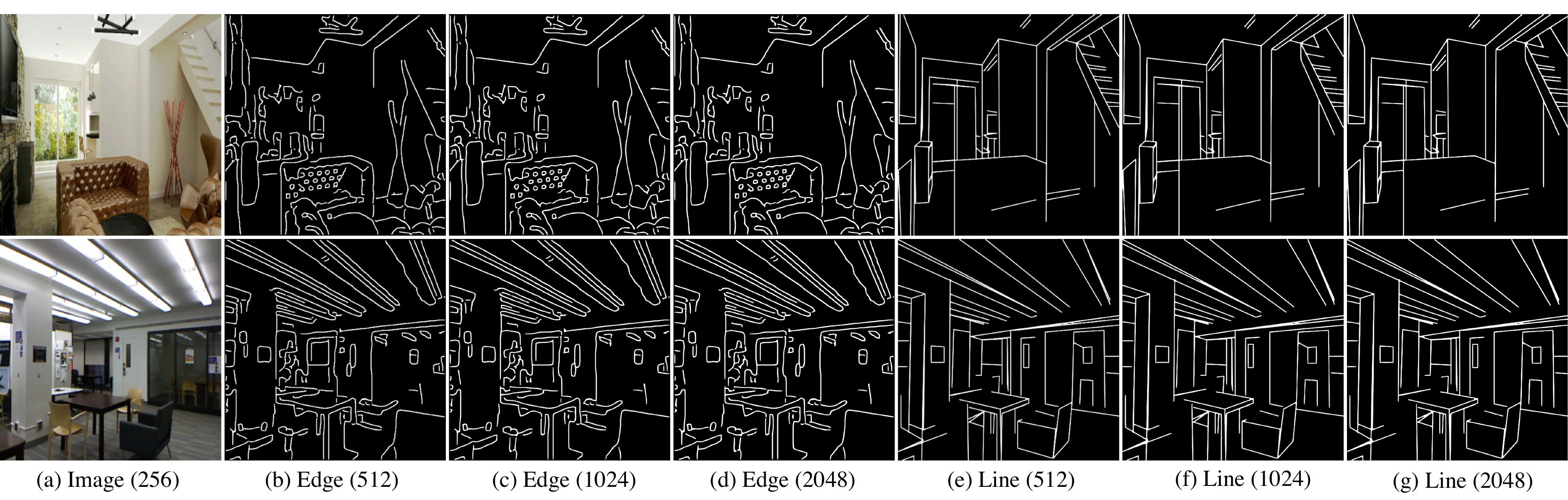}
\par\end{centering}
\caption{Iteratively outputs of SSU which have sizes from 512 to 2048. The results are consistent and robust.}
\label{fig:SSU_iterative}
\end{figure*}

Our Simple Structure Upsampler (SSU) introduced in Sec 3.2 can also work iteratively for larger image sizes. First, we should process the output edges $\mathbf{I}'_e$ and lines $\mathbf{I}'_l$ of SSU through shifted sigmoid as
\begin{equation}
\begin{split}
\mathbf{I}'_e&=\mathrm{sigmoid}[\gamma(\mathbf{I}'_e+\beta)],\\
\mathbf{I}'_l&=\mathrm{sigmoid}[\gamma(\mathbf{I}'_l+\beta)],
\end{split}
\label{eq:ssu_output}
\end{equation}
where $\gamma=2, \beta=2$ in our evaluation, and $\gamma,\beta$ are randomly selected from $[1.5, 3]$ for the finetuning. Since the output size of SSU is doubled, we can repeat the inputs $\mathbf{I}_e,\mathbf{I}_l\in\mathbb{R}^{h\times w}$ for $q$ times to achieve $\mathbf{I}'_e,\mathbf{I}'_l\in\mathbb{R}^{2^qh\times 2^qw}$. Then, the outputs can further be resized with the bilinear interpolation for the target size. In general, our SSU can get good and robust upsampled results for large sizes as shown in Fig.~\ref{fig:SSU_iterative}.

\section{Supplementary Experiments}
In this section, we provide some more qualitative and quantitative results to show the effects of our proposed components. Moreover, some details about the post-processing are also discussed.

\subsection{More Qualitative Results}

\begin{figure*}
\begin{centering}
\includegraphics[width=0.75\linewidth]{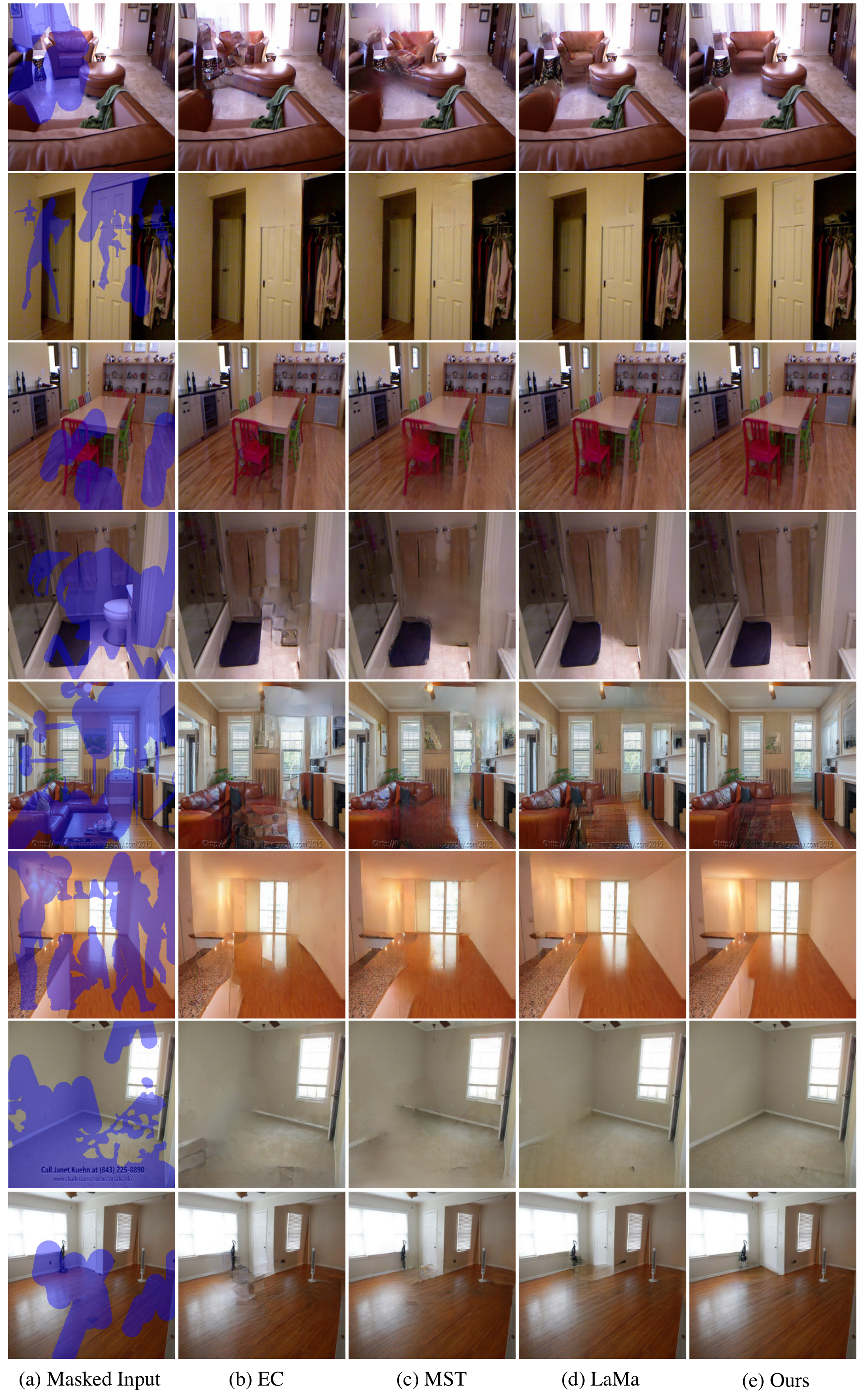}
\par\end{centering}
\caption{Qualitative results of Indoor dataset compared among EC~\cite{nazeri2019edgeconnect}, MST~\cite{cao2021learning}, LaMa~\cite{suvorov2021resolution}, and ours. Zoom-in for details}
\label{fig:indoor_quali}
\end{figure*}

\begin{figure*}
\begin{centering}
\includegraphics[width=0.99\linewidth]{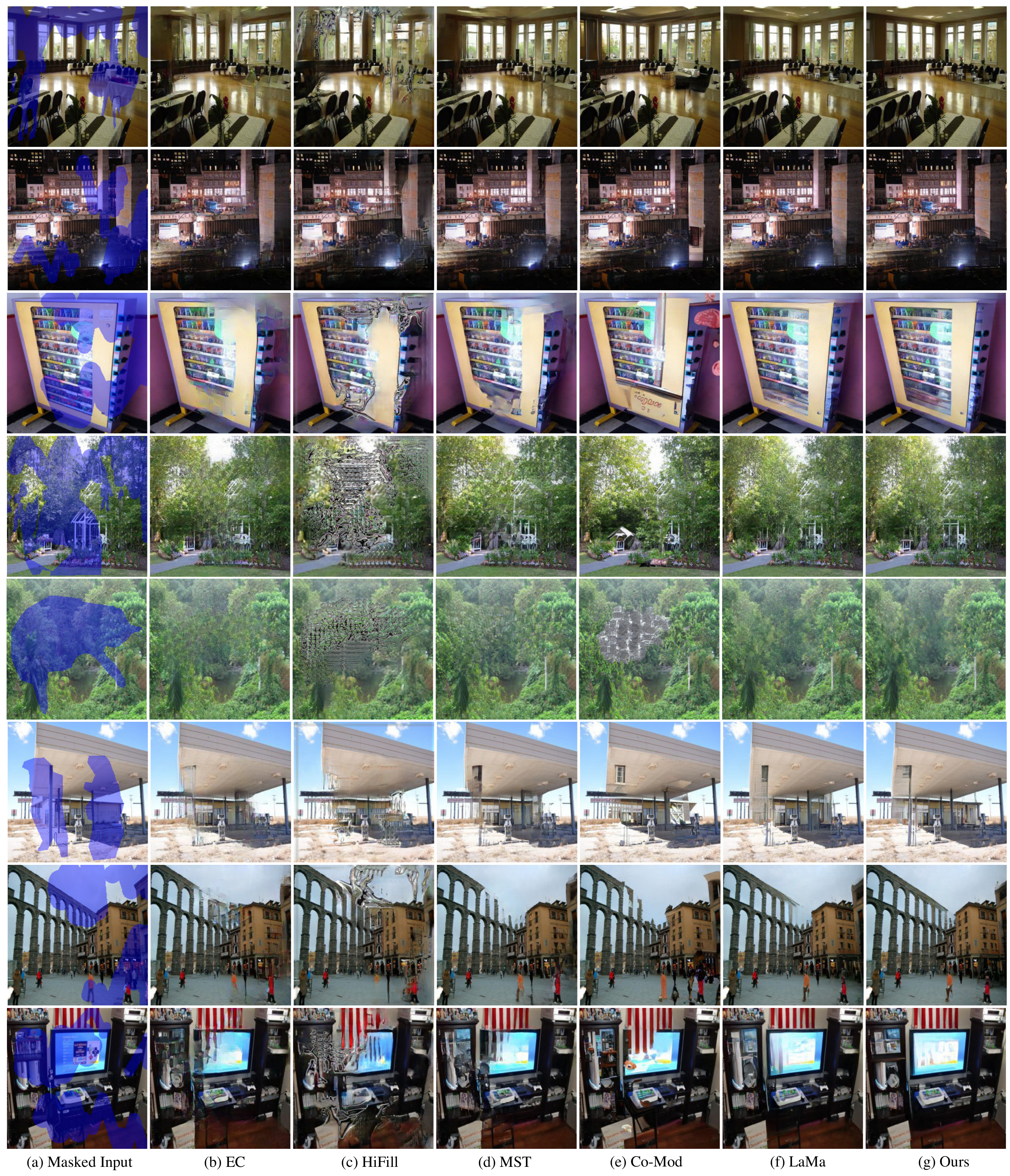}
\par\end{centering}
\caption{Qualitative results of Places2 dataset compared among EC~\cite{nazeri2019edgeconnect}, HiFill~\cite{yi2020contextual}, MST~\cite{cao2021learning}, Co-Mod~\cite{zhao2021large}, LaMa~\cite{suvorov2021resolution}, and ours. Zoom-in for details}
\label{fig:places2_quali}
\end{figure*}

More qualitative results of Indoor and Places2 are shown in Fig.~\ref{fig:indoor_quali} and Fig.~\ref{fig:places2_quali}. Note that our method not only achieves better results in many man-made scenes, but also gets competitive results in natural scenes benefited from MPE and edges.

\subsection{Quantitative Results with Different Masks}

\begin{table*}
\caption{Quantitative inpainting results on Indoor and Places2 with different mask ratios. \label{table:main_results}}
\vspace{-0.15in}
\begin{centering}
\small
\begin{tabular}{c|c|cccc|cccccc}
\hline 
 &  & \multicolumn{4}{c|}{Indoor (256$\times$256)} & \multicolumn{6}{c}{Places2 (256$\times$256)}\tabularnewline
\hline 
\hline 
 & Mask & EC & MST & LaMa & Ours & EC & HiFill & Co-Mod & MST & LaMa & Ours\tabularnewline
\hline 
\multirow{5}{*}{PSNR$\uparrow$} & 10\textasciitilde 20\% & 28.18 & 28.72 & 29.05 & \textbf{29.87} & 26.60 & 24.04 & 26.40 & 28.13 & 28.23 & \textbf{28.31}\tabularnewline
 & 20\textasciitilde 30\% & 25.14 & 25.66 & 25.96 & \textbf{26.66} & 24.26 & 21.64 & 23.61 & 25.07 & 25.31 & \textbf{25.40}\tabularnewline
 & 30\textasciitilde 40\% & 23.02 & 23.53 & 23.87 & \textbf{24.64} & 22.59 & 19.96 & 21.73 & 23.07 & 23.43 & \textbf{23.51}\tabularnewline
 & 40\textasciitilde 50\% & 21.55 & 22.02 & 22.39 & \textbf{23.13} & 21.27 & 18.63 & 20.28 & 21.53 & 22.03 & \textbf{22.11}\tabularnewline
 & Mixed & 24.07 & 24.52 & 25.20 & \textbf{25.57} & 23.31 & 20.76 & 22.57 & 24.02 & 24.37 & \textbf{24.42}\tabularnewline
\hline 
\multirow{5}{*}{SSIM$\text{\ensuremath{\uparrow}}$} & 10\textasciitilde 20\% & 0.951 & 0.954 & 0.956 & \textbf{0.961} & 0.913 & 0.883 & 0.926 & 0.941 & \textbf{0.942} & \textbf{0.942}\tabularnewline
 & 20\textasciitilde 30\% & 0.916 & 0.922 & 0.925 & \textbf{0.933} & 0.872 & 0.818 & 0.880 & 0.898 & 0.901 & \textbf{0.902}\tabularnewline
 & 30\textasciitilde 40\% & 0.876 & 0.886 & 0.890 & \textbf{0.901} & 0.828 & 0.751 & 0.831 & 0.852 & 0.859 & \textbf{0.860}\tabularnewline
 & 40\textasciitilde 50\% & 0.835 & 0.848 & 0.855 & \textbf{0.870} & 0.783 & 0.682 & 0.781 & 0.803 & 0.814 & \textbf{0.817}\tabularnewline
 & Mixed & 0.884 & 0.894 & 0.902 & \textbf{0.907} & 0.839 & 0.770 & 0.843 & 0.862 & 0.869 & \textbf{0.870}\tabularnewline
\hline 
\multirow{5}{*}{FID$\downarrow$} & 10\textasciitilde 20\% & 9.56 & 8.56 & 8.01 & \textbf{7.18} & 1.95 & 4.71 & 0.52 & 0.76 & 0.45 & \textbf{0.43}\tabularnewline
 & 20\textasciitilde 30\% & 16.22 & 15.88 & 13.23 & \textbf{12.13} & 3.79 & 11.93 & 1.00 & 1.86 & 0.95 & \textbf{0.88}\tabularnewline
 & 30\textasciitilde 40\% & 23.48 & 22.69 & 18.77 & \textbf{16.51} & 6.98 & 25.16 & 1.64 & 3.83 & 1.72 & \textbf{1.55}\tabularnewline
 & 40\textasciitilde 50\% & 31.16 & 31.06 & 23.47 & \textbf{20.87} & 11.49 & 44.68 & \textbf{2.38} & 6.80 & 2.81 & 2.51\tabularnewline
 & Mixed & 22.02 & 21.65 & 16.97 & \textbf{15.93} & 6.21 & 21.33 & 1.49 & 3.53 & 1.63 & \textbf{1.47}\tabularnewline
\hline 
\multirow{5}{*}{LPIPS$\downarrow$} & 10\textasciitilde 20\% & 0.054 & 0.050 & 0.044 & \textbf{0.038} & 0.073 & 0.119 & 0.053 & 0.047 & 0.047 & \textbf{0.042}\tabularnewline
 & 20\textasciitilde 30\% & 0.094 & 0.087 & 0.078 & \textbf{0.068} & 0.111 & 0.189 & 0.098 & 0.082 & 0.083 & \textbf{0.073}\tabularnewline
 & 30\textasciitilde 40\% & 0.140 & 0.129 & 0.117 & \textbf{0.101} & 0.152 & 0.265 & 0.140 & 0.120 & 0.121 & \textbf{0.107}\tabularnewline
 & 40\textasciitilde 50\% & 0.189 & 0.172 & 0.156 & \textbf{0.136} & 0.194 & 0.343 & 0.184 & 0.160 & 0.161 & \textbf{0.143}\tabularnewline
 & Mixed & 0.135 & 0.122 & 0.112 & \textbf{0.098} & 0.149 & 0.137 & 0.246 & 0.122 & 0.155 & \textbf{0.108}\tabularnewline
 \hline
\end{tabular}
\par\end{centering}
\vspace{-0.1in}
\end{table*}

We show more quantitative results with different masking rates from 10\% to 50\% and mixture of segmentation and irregular masks in Tab.~\ref{table:main_results}.

\subsection{More Structural Experiments}

\noindent\textbf{TSR Ablations.} 
For the Indoor dataset, we conducted several ablation experiments on our Transformer Structure Restoration (TSR), and the results are displayed in Tab.~\ref{table:axial_memory_speed} and  Tab.~\ref{table:abla_TSR}. As illustrated in Tab.~\ref{table:axial_memory_speed} and the first two rows of Tab.~\ref{table:abla_TSR}, replacing one standard self-attention module~\cite{vaswani2017attention} with an axial attention module~\cite{ho2019axial} in our Transformer Block can greatly reduce the GPU memory usage and speed up the model inference while keeping all metrics basically unchanged. Furthermore, we add the relative position encoding (RPE)~\cite{raffel2019exploring} into our axial attention module, which can boost our results. Note that the RPE must be incorporated with the axial attention module in row-wise and column-wise, while standard attention based RPE costs much more GPU memory due to the long sequence. On the other hand, as we think that a higher recall will benefit the later image inpainting, we further multiply the line logits by 4 before feeding it through the sigmoid activation function in all the experiments. This strategy enhances recall while only compromising a little precision.

\begin{table}
\small
\caption{Efficient ablations of axial attention module. FPS is the Frames Per Second during the inference. The GPU memory is test on single Tesla V100 with batch size 8.\label{table:axial_memory_speed}}
\begin{centering}
\begin{tabular}{ccc}
\hline
 & FPS & GPU Memory (MB)\tabularnewline
\hline
\hline
w./o. Axial & 6.41 & 14845\tabularnewline
with Axial & \textbf{7.89} & \textbf{10547}\tabularnewline
\hline
\end{tabular}
\par\end{centering}
\end{table}

\begin{table*}
\small
\caption{Ablation studies of TSR on the Indoor dataset, where P., R., F1 mean Precision, Recall, and F1-score.\label{table:abla_TSR}}
\begin{centering}
\begin{tabular}{cc|cccccc|c}
\hline
 &  & \multicolumn{3}{c|}{Edge} & \multicolumn{3}{c|}{Line} & Avg\tabularnewline
Axial & RPE & P. & R. & \multicolumn{1}{c|}{F1} & P. & R. & F1 & F1\tabularnewline
\hline
\hline
 &  & 38.27 & 33.12 & 34.78 & 52.93 & 65.79 & 57.73 & 46.26\tabularnewline
\CheckmarkBold{} &  & \textbf{38.30} & 32.90 & 34.64 & 52.74 & \textbf{66.48} & 57.87 & 46.26\tabularnewline
\CheckmarkBold{} & \CheckmarkBold{} & 37.34 & \textbf{34.25} & \textbf{35.10} & \textbf{53.60} & 66.23 & \textbf{58.35} & \textbf{46.72}\tabularnewline
\hline
\end{tabular}
\par\end{centering}
\end{table*}

\begin{figure*}
\begin{centering}
\includegraphics[width=0.88\linewidth]{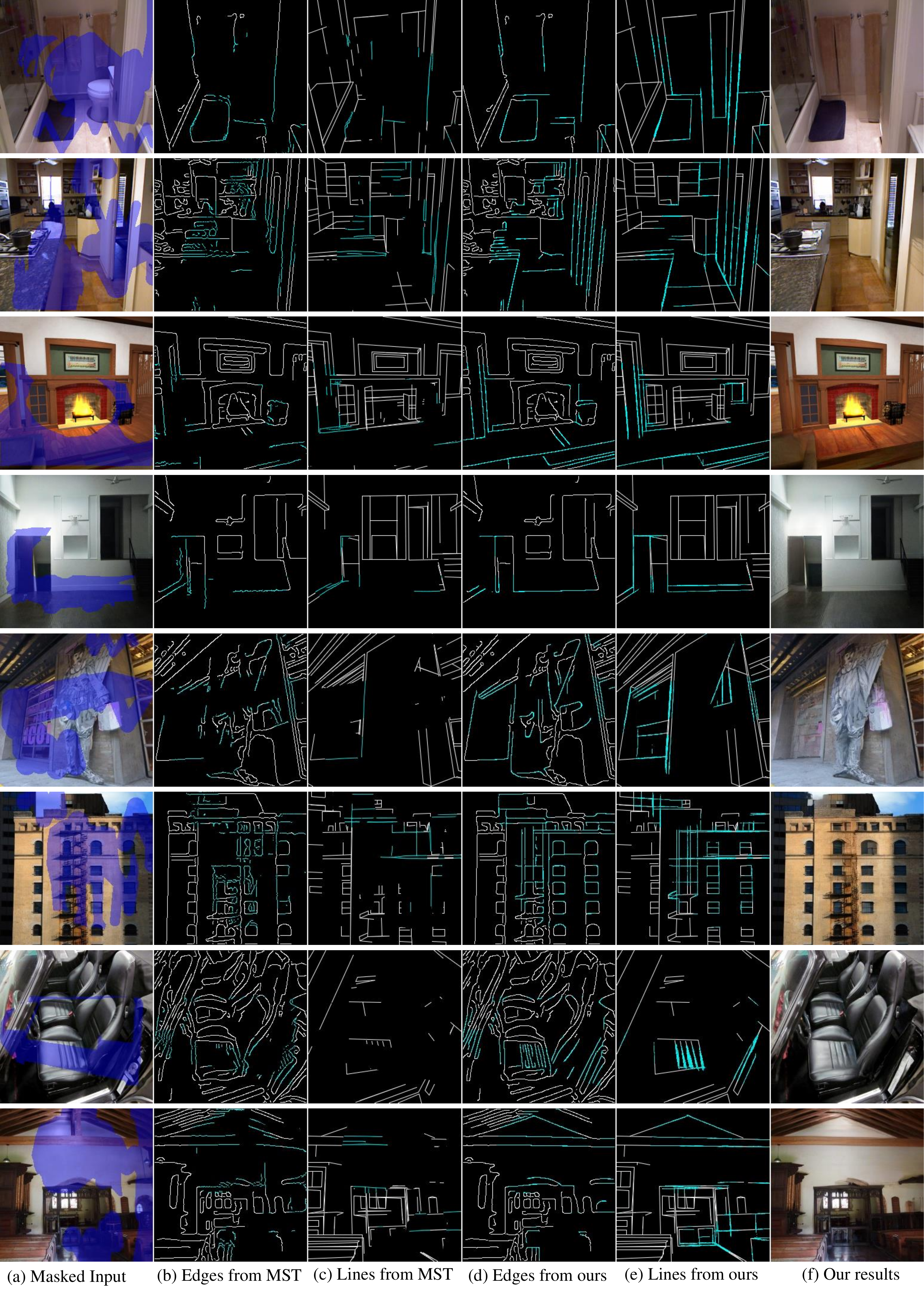}
\par\end{centering}
\caption{Predicted edges, lines and inpainted results in Indoor and Places2 compared with MST~\cite{cao2021learning}. The first four examples are from the Indoor dataset; the last four examples are from the Places2 dataset. Blue edges (lines) indicate reconstruction from models.}
\label{fig:stru_comp}
\end{figure*}

\noindent\textbf{More Structural Qualitative Results.} We show some more structural results of TSR in Fig.~\ref{fig:stru_comp} compared with MST~\cite{cao2021learning}. Our TSR can outperform the CNN based method. Furthermore, edges and lines from TSR can effectively guide the final inpainted results.

\subsection{Effects of Mask-Predict}
During the inference, Mask-Predict~\cite{cho2020xlxmert, guo2020incorporating, ghazvininejad2019mask} is used in TSR which is a non-autoregressive sampling method. Mask-Predict predicts all target pixels at the first iteration. Then we re-mask and re-predict pixels with low-confidence iteratively. Mask-Predict can greatly enrich the generated results without heavy costs. 

\begin{figure*}
\begin{centering}
\includegraphics[width=0.99\linewidth]{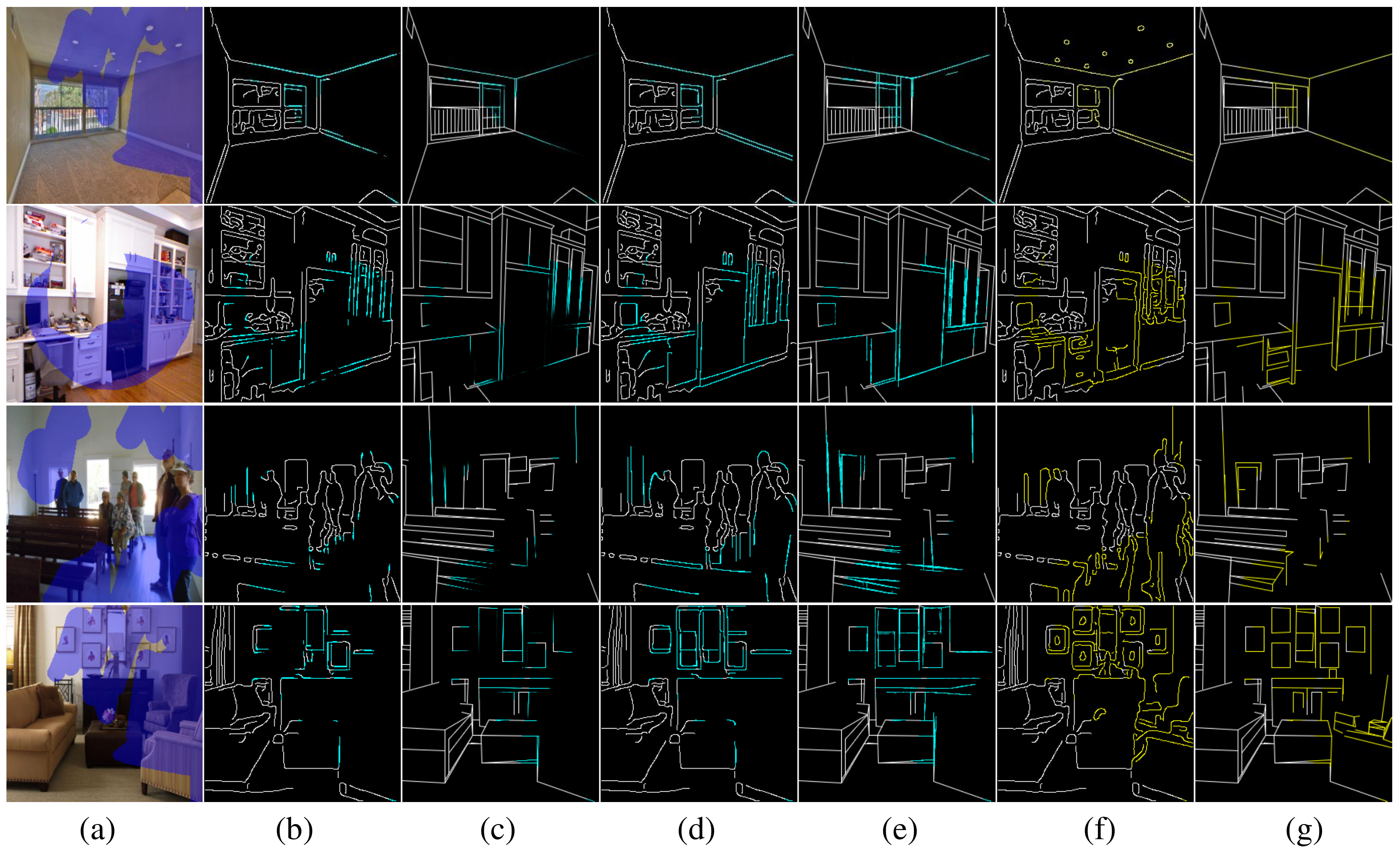}
\par\end{centering}
\caption{Ablation studies on Mask-Predict. From left to right: (a) Image, (b) Reconstructed edge w/o Mask-Predict, (c) Reconstructed line w/o Mask-Predict, (d) Reconstructed edge with Mask-Predict, (e) Reconstructed line with Mask-Predict, (f) Ground truth edge, (g) Ground truth line. The first two rows are from the Indoor dataset; the last two rows are from the Places2 dataset.  Blue and yellow edges (lines) indicate reconstruction and ground truth within mask region respectively.}
\label{fig:ablation_stru}
\end{figure*}

\begin{figure*}
\begin{centering}
\includegraphics[width=0.99\linewidth]{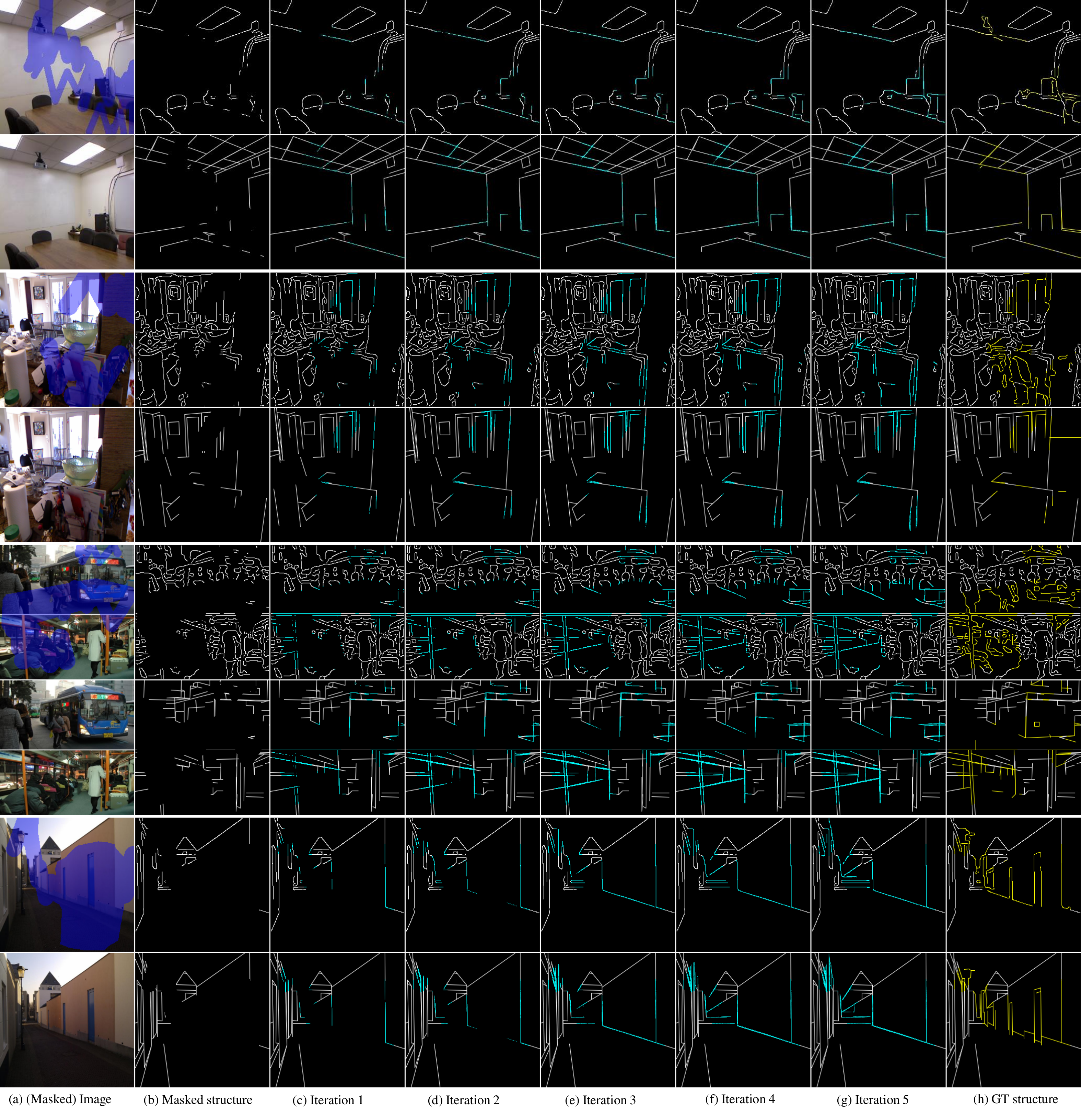}
\par\end{centering}
\caption{Mask-Predict for edges and lines in Indoor. The first two examples are from the Indoor dataset; the last two examples are from the Places2 dataset. For each example, the structure in the first row is the edge; the structure in the second row is the line. Blue and yellow edges (lines) indicate our reconstruction and ground truth within mask region respectively.}
\label{fig:sampling}
\end{figure*}

Since our TSR can output a $256\times 256$ probability map for edge and line respectively; of course, we can directly use this probability map as the repair result of our edge and line. However, we find that the recall is still insufficient.
Fig.~\ref{fig:ablation_stru}(b)(c) show that it can only predict a few regions with high confidence. But the probability confidence for the inner masked region is low, which leads to incomplete structures. As a result, we employ the Mask-Predict~~\cite{cho2020xlxmert, guo2020incorporating, ghazvininejad2019mask} strategy. It predicts all target pixels at the first iteration. Then we re-mask and re-predict pixels with low-confidence iteratively for a constant number of iterations. Note that we just re-mask edge and line without re-masking the input masked image.  This technique can achieve more complete structural information as illustrated in Fig.~\ref{fig:ablation_stru}(d)(e). We set the total number of iterations to 5 in our experiment for a trade-off between inference time and recall. Fig.~\ref{fig:sampling} shows our outcomes with different Mask-Predict iterations. Note that we ignore pixels with low confidence for each iteration in Fig.~\ref{fig:sampling}, so the iteration 1 results of Fig.~\ref{fig:sampling}(c) look different from the results without Mask-Predict.
The inside portion of the mask can be gradually restored with larger iterations as shown in Fig.~\ref{fig:sampling}. 

\subsection{User Study}

\begin{figure}
\begin{centering}
\includegraphics[width=0.99\linewidth]{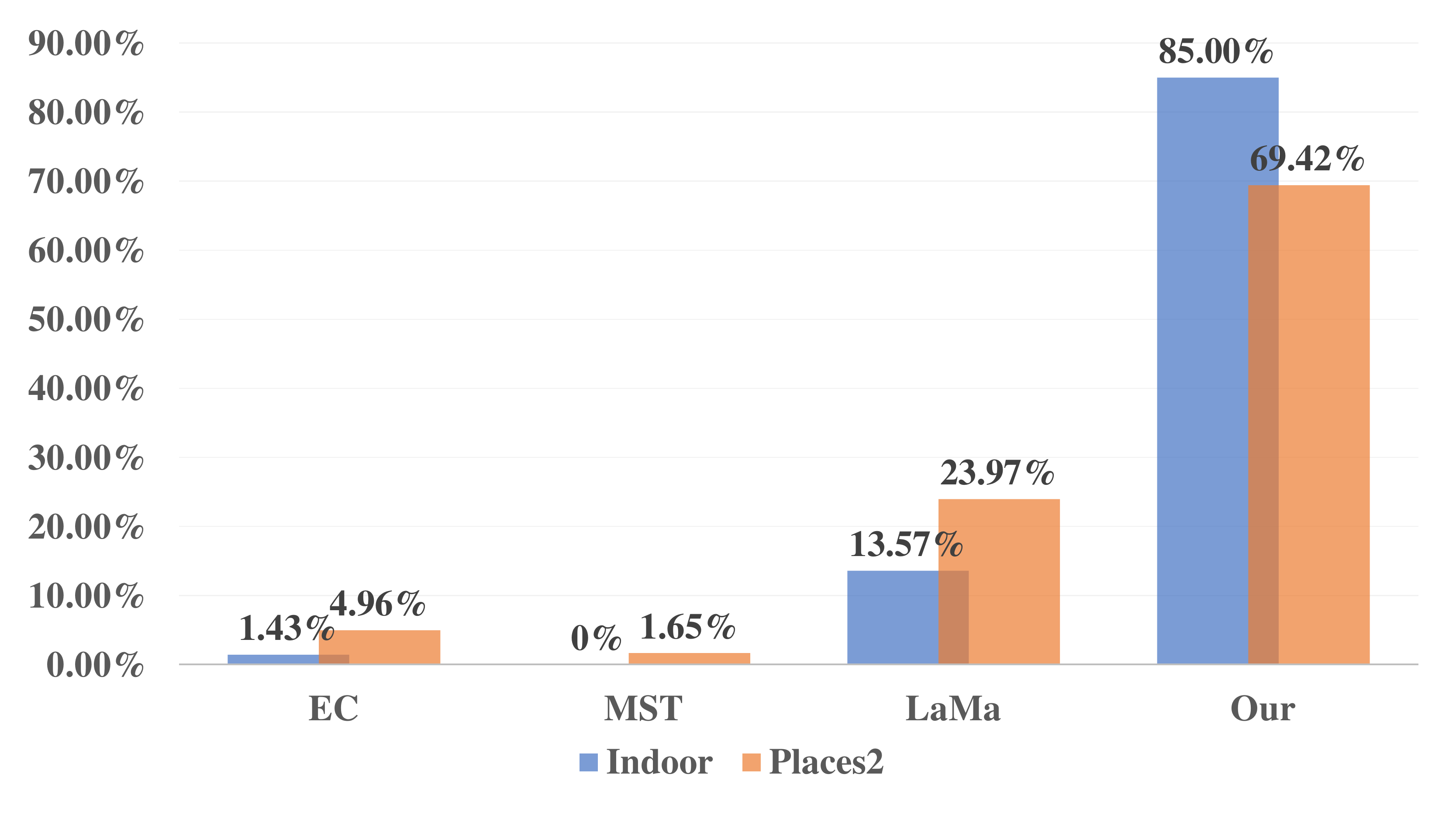}
\par\end{centering}
\caption{Average scores of Indoor and Places2 for user studies, which are
collected from volunteers who select the best one from shuffled inpainted images.}
\label{fig:user_study}
\end{figure}

We conduct user study on several models to validate the effectiveness of our model from the perspective of human. Specifically, we invite 10 volunteers who are not familiar with image inpainting to judge the quality of inpainted images.  On Indoor and Places2, four methods are compared, which including EC~\cite{nazeri2019edgeconnect}, MST~\cite{cao2021learning} LaMa~\cite{suvorov2021resolution} and ours. Given the masked inputs, we randomly shuffle and combine the results of four methods together. Then, volunteers are required to choose the best one from each group. As shown in Fig.~\ref{fig:user_study}, our method outperforms other three competitors on both two datasets. Especially, our method can achieve a great advantage compared with the baseline method \emph{i.e.}, LaMa.

\subsection{Results of Rectangular Masks}

\begin{figure}
\begin{centering}
\includegraphics[width=0.99\linewidth]{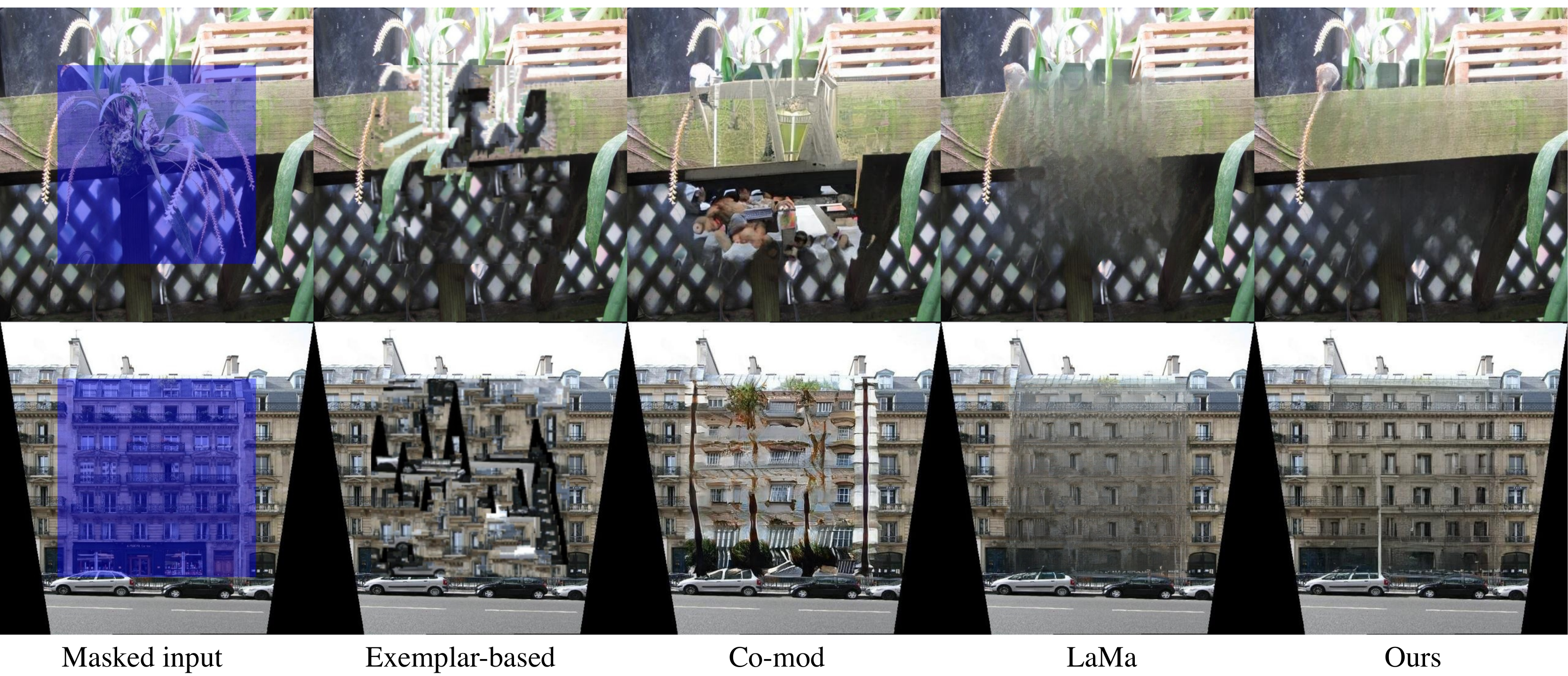}
\par\end{centering}
\caption{Inpainting results of 512 images compared with Exemplar-based inpainting~\cite{criminisi2003object}, Co-Mod, LaMa, and ours.}
\label{fig:rec}
\end{figure}

Here we provide some results of 40\% center rectangular masks of 1k Places(512) images without any retraining in Tab.~\ref{table:rectangular_mask}. Note that Co-Mod~\cite{zhao2021large} is the only one trained with some rectangular masks while other methods have not been trained with similar masks. Moreover, we compare related qualitative results in Fig.~\ref{fig:rec}. And the classical exemplar-based inpainting~\cite{criminisi2003object} is also included. Traditional exemplar-based method fails to work properly and is time-consuming. Co-Mod has hallucinated artifacts instead of generating plausible results. And LaMa results are blur with still high PSNR.

\begin{table}
\small
\caption{Quantitative results on 1k Places 512 images with 40\% center rectangular masks.\label{table:rectangular_mask}}
\begin{centering}
\begin{tabular}{cccc}
\hline 
 & {PSNR} & {FID} & {LPIPS}\tabularnewline
\hline 
{Co-Mod} & {17.59} & \textbf{52.38} & {0.262}\tabularnewline
{LaMa} & \textbf{19.69} & {61.67} & {0.268}\tabularnewline
{Ours} & {19.65} & {55.85} & \textbf{0.239}\tabularnewline
\hline 
\end{tabular}
\par\end{centering}
\end{table}

\subsection{Comparisons of Texture Images}

\begin{table}
\small
\caption{Quantitative results on 512 texture images from~\cite{cimpoi2014describing}.\label{table:texture}}
\begin{centering}
\begin{tabular}{ccc}
\hline 
 & {LaMa} & {Ours}\tabularnewline
\hline 
{PSNR} & \textbf{25.82} & {25.67}\tabularnewline
{SSIM} & \textbf{0.875} & {0.869}\tabularnewline
{FID} & {12.86} & \textbf{11.67}\tabularnewline
{LPIPS} & {0.138} & \textbf{0.134}\tabularnewline
\hline 
\end{tabular}
\par\end{centering}
\end{table}

\begin{figure}
\begin{centering}
\includegraphics[width=0.99\linewidth]{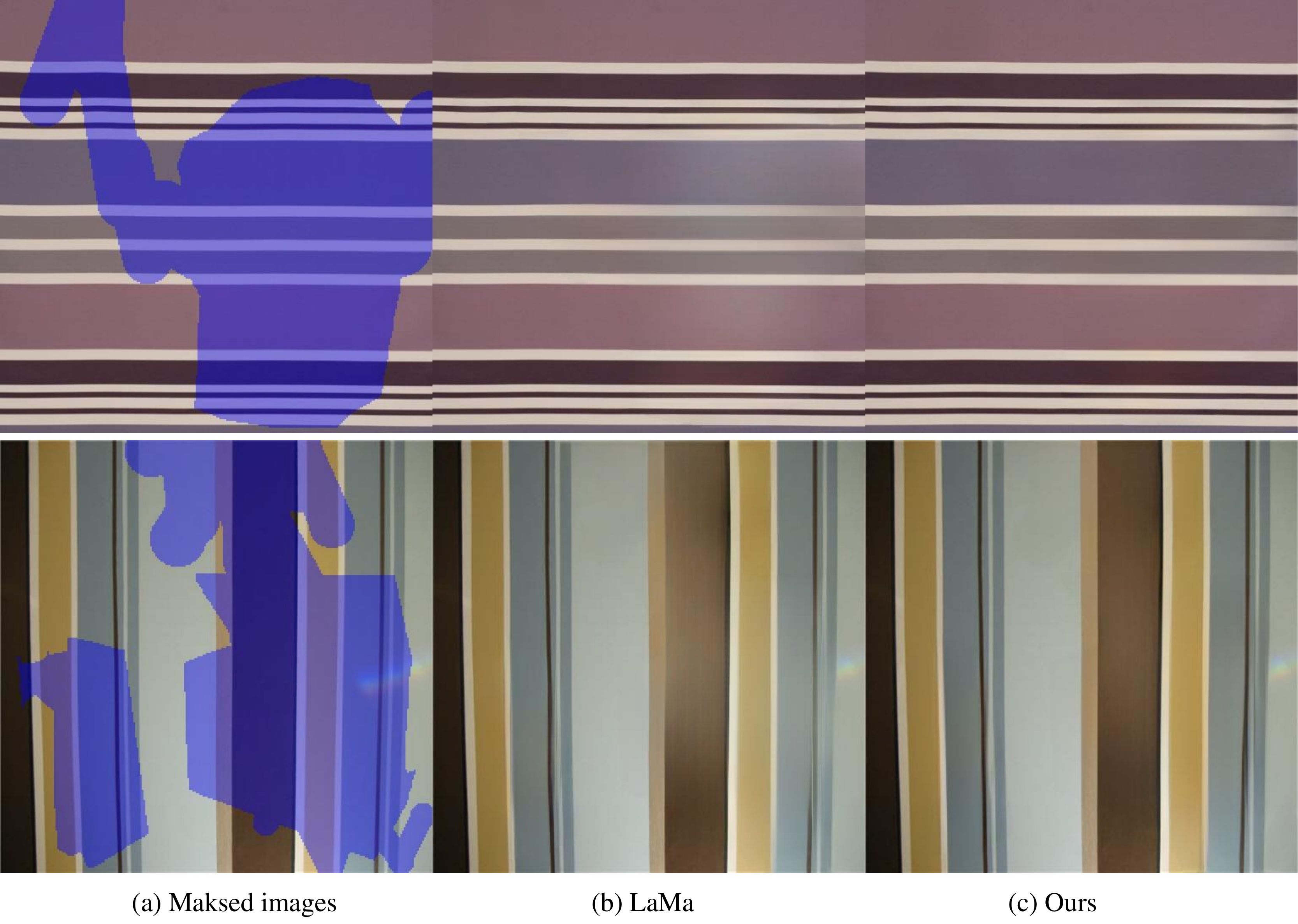}
\par\end{centering}
\caption{Inpainting results of 512 texture images~\cite{cimpoi2014describing} compared with LaMa and ours.}
\label{fig:texture}
\end{figure}

We further compare our method with LaMa on 1,880 texture images~\cite{cimpoi2014describing} in Tab.~\ref{table:texture} and Fig.~\ref{fig:texture}, which contain strong periodic textures. Since this dataset is very suitable to LaMa~\cite{suvorov2021resolution}, our method still has competitive performance.

\begin{figure}
\begin{centering}
\includegraphics[width=0.95\linewidth]{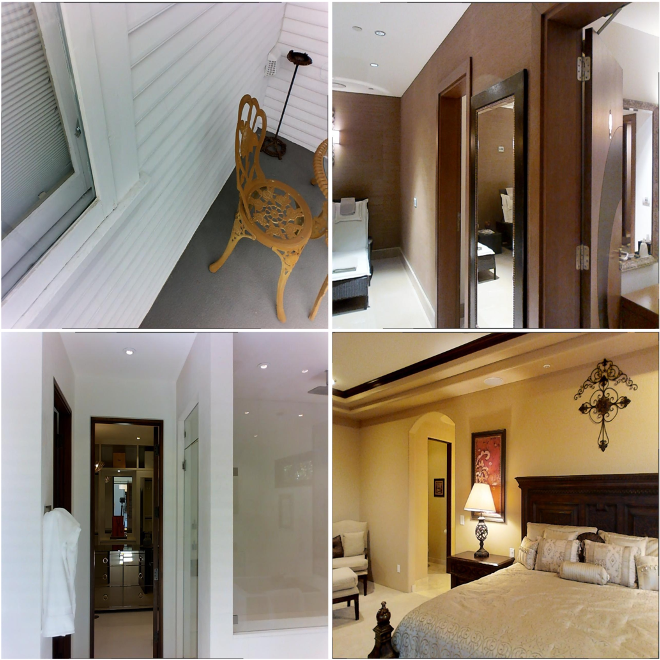}
\par\end{centering}
\caption{Examples of resized 1024$\times$1024 MatterPort3D images.}
\label{fig:matterport3d}
\end{figure}

\begin{figure}
\begin{centering}
\includegraphics[width=0.95\linewidth]{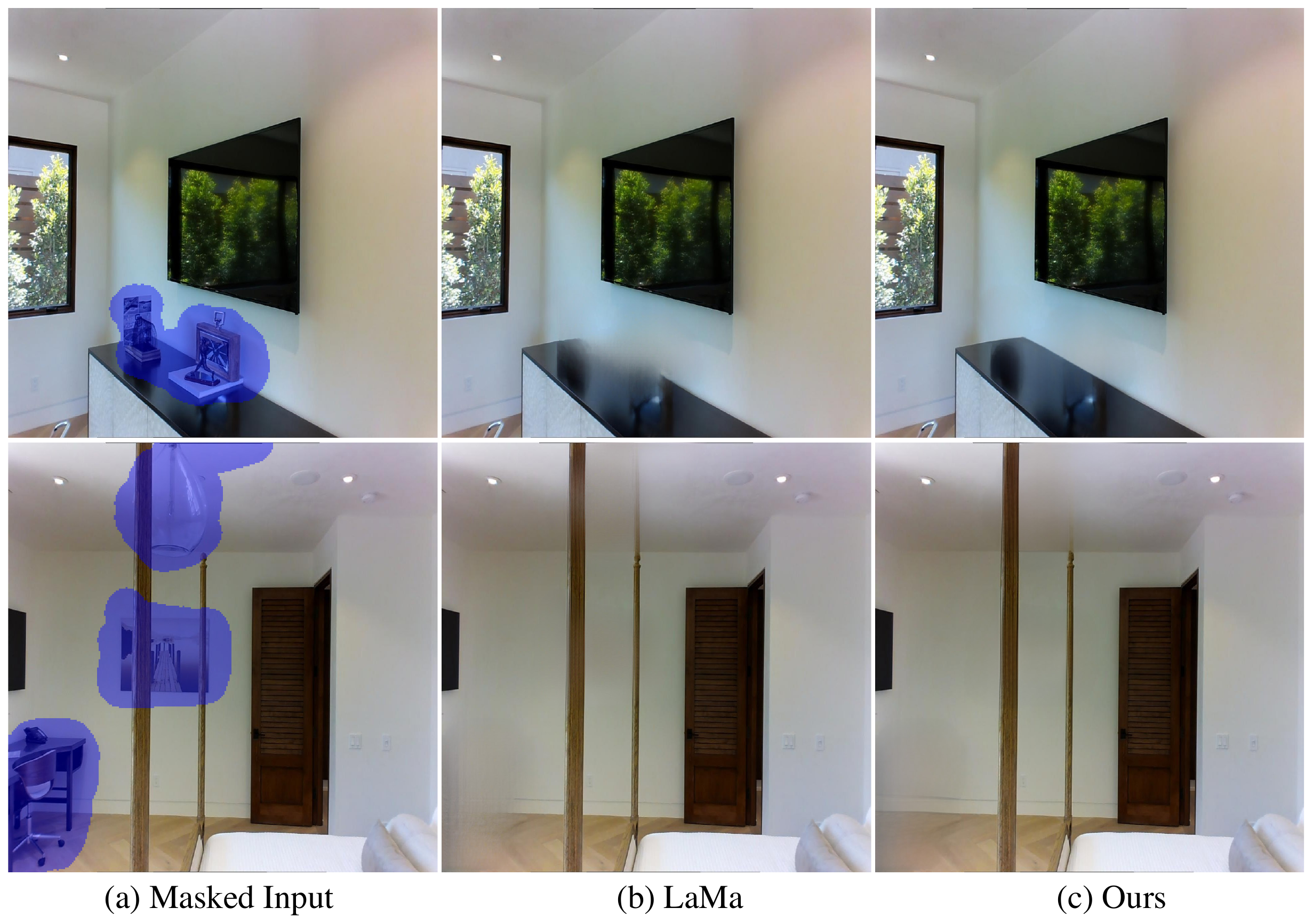}
\par\end{centering}
\caption{Inpainting results of LaMa~\cite{suvorov2021resolution} and ours in 1024$\times$1024 MatterPort3D images.}
\label{fig:matterport3d_res}
\end{figure}

\subsection{Results of MatterPort3D}
We use the test set of MatterPort3D~\cite{chang2017matterport3d} to evaluate the effectiveness of our method in the high-resolution structure recovery. MatterPort3D images tested in this paper are consisted of 1,965 indoor images in 1280$\times$1024. We resized them into 1024$\times$1024 as shown in Fig.~\ref{fig:matterport3d}. We provide some qualitative results of our method and LaMa compared on MatterPort3D in Fig.~\ref{fig:matterport3d_res}. For these structural images, our results enjoy better structures.

\section{More High Resolution Results}

\begin{figure*}
\begin{centering}
\includegraphics[width=0.99\linewidth]{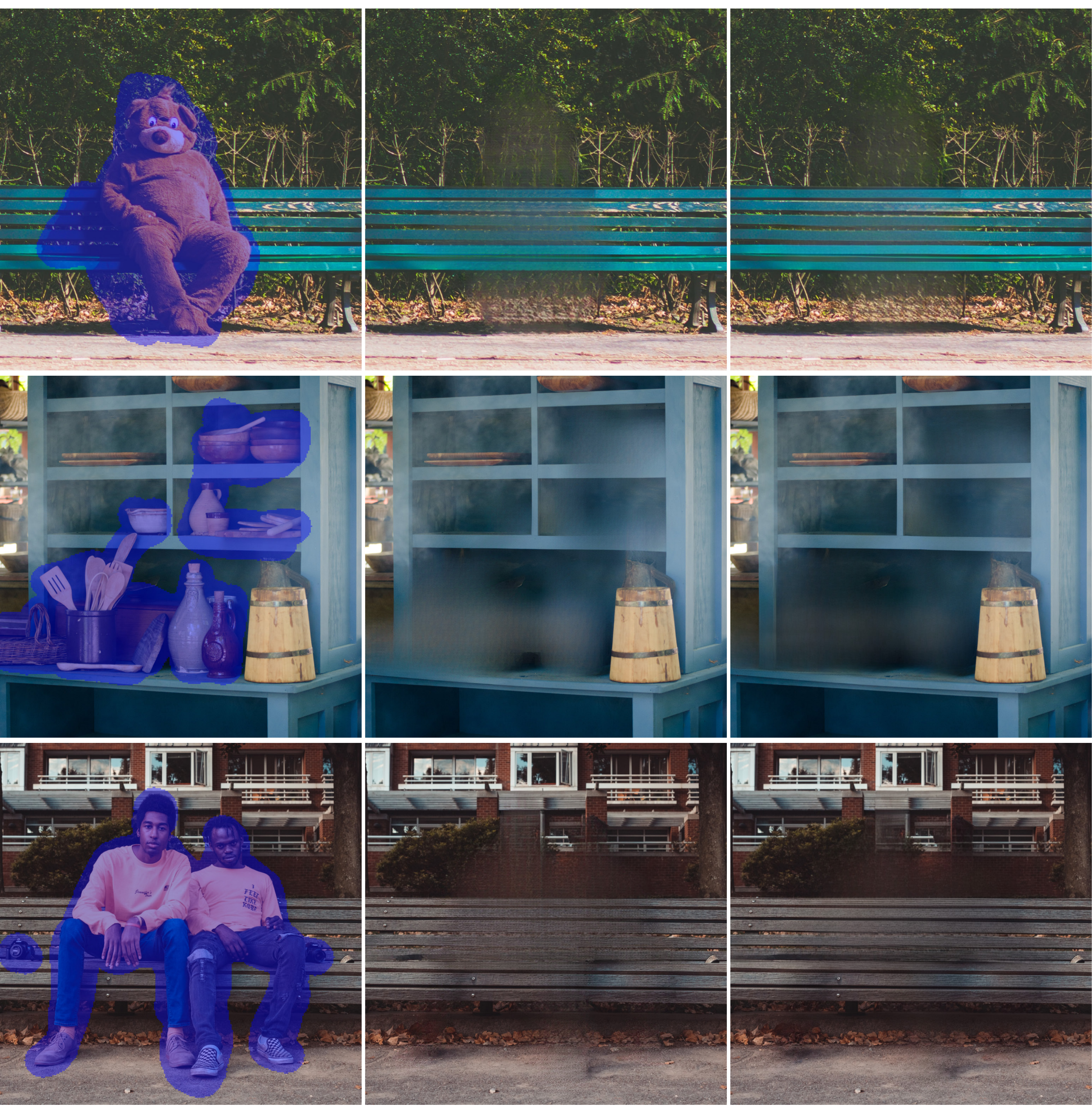}
\par\end{centering}
\caption{High-resolution object removal results. From left to right are masked inputs, results from LaMa~\cite{suvorov2021resolution}, results from our method. Please zoom-in for more details.}
\label{fig:high_res1}
\end{figure*}

\begin{figure*}
\begin{centering}
\includegraphics[width=0.99\linewidth]{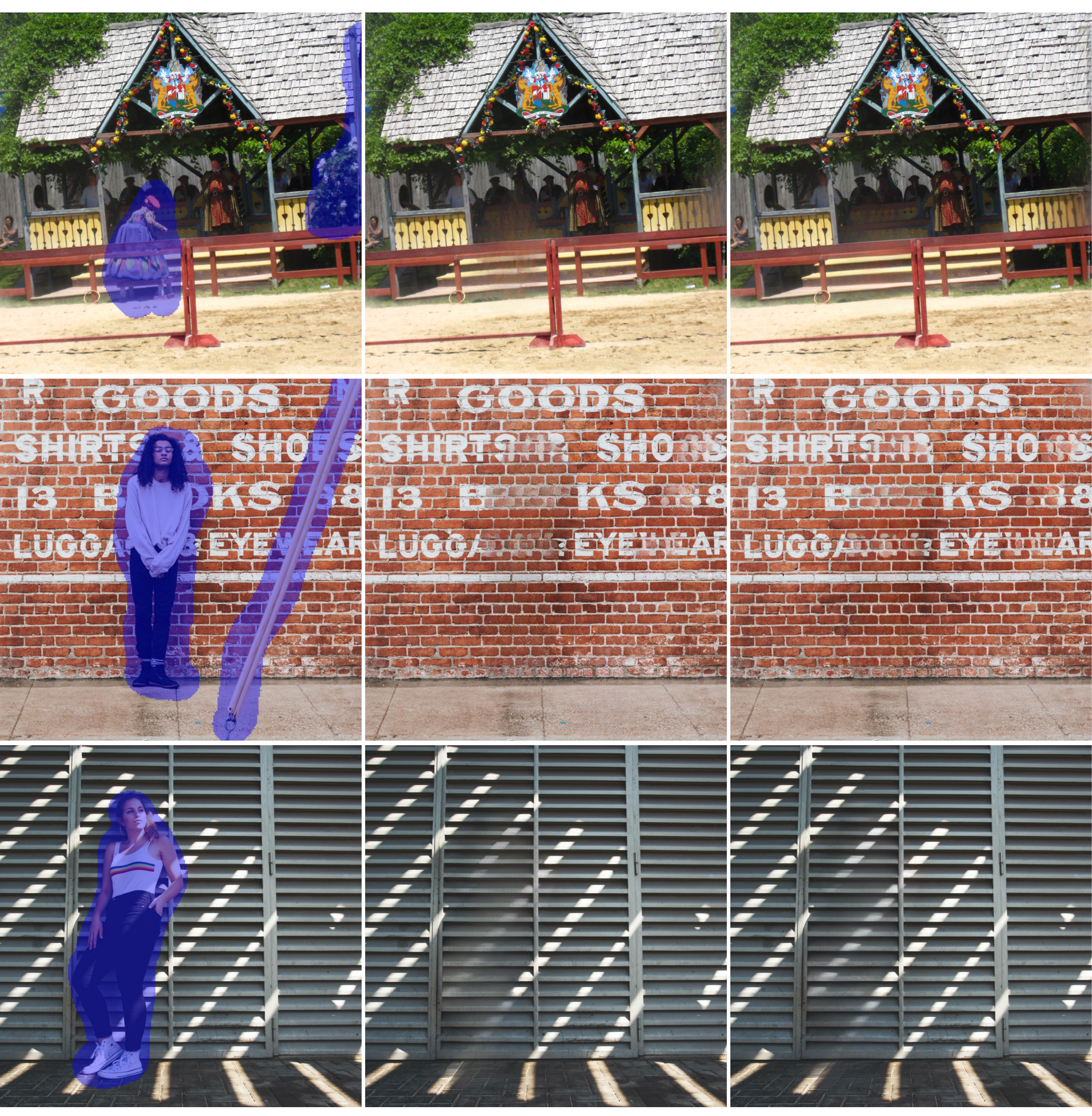}
\par\end{centering}
\caption{High-resolution object removal results. From left to right are masked inputs, results from LaMa~\cite{suvorov2021resolution}, results from our method. Please zoom-in for more details.}
\label{fig:high_res2}
\end{figure*}

\begin{figure*}
\begin{centering}
\includegraphics[width=0.99\linewidth]{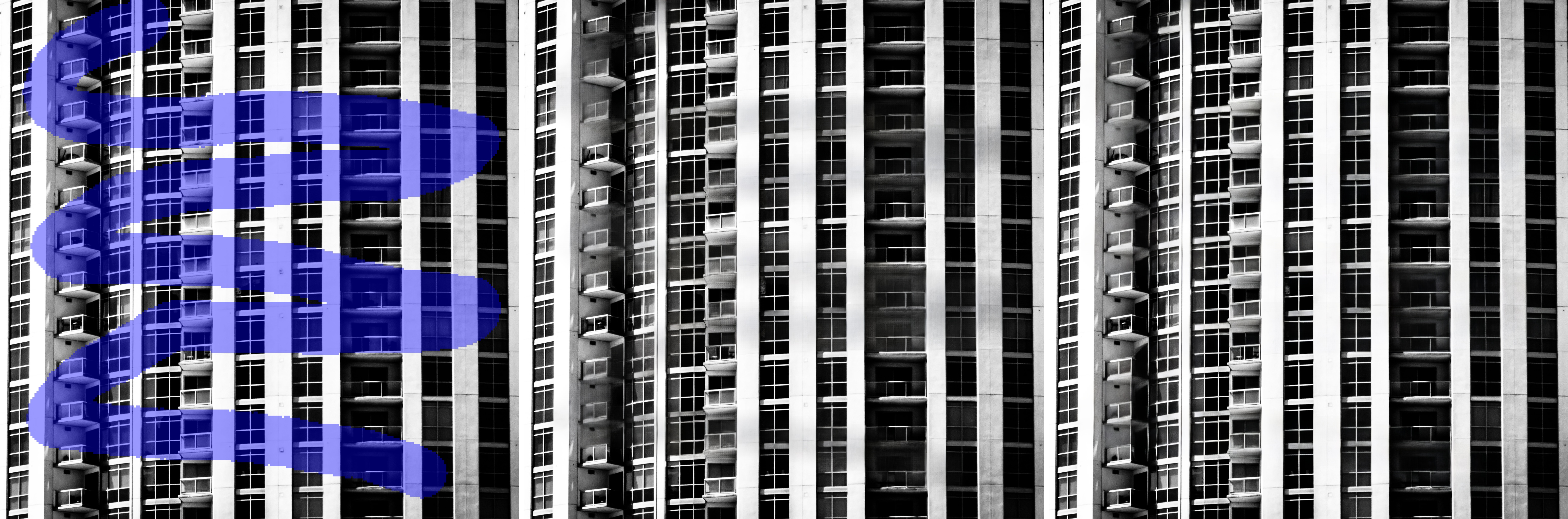}
\par\end{centering}
\caption{The high-resolution inpainting comparison of 2048$\times$2048. From left to right are the masked input, the result from LaMa~\cite{suvorov2021resolution}, the result from our method. Please zoom-in for more details.}
\label{fig:high_res3}
\end{figure*}

In Fig.~\ref{fig:high_res1}, Fig.~\ref{fig:high_res2}, and Fig.~\ref{fig:high_res3} we provide some object removal instances in large images from 1k to 2k resolutions compared with LaMa~\cite{suvorov2021resolution}. Some cases are selected from the open-source testset of LaMa. Note that our method outperforms LaMa in scenes with weak textures such as row 2 in Fig.~\ref{fig:high_res1} and row 1 in Fig.~\ref{fig:high_res2}. For the cases with sparse regular textures and lines (rows 1,3 of Fig.~\ref{fig:high_res1}), our method can still achieve more clear borderlines. For the cases with dense regular textures (rows 2,3 of Fig.~\ref{fig:high_res2}), LaMa gets competitive results, which shows that FFC in frequency fields has solved these problems properly. However, our method can also achieve results with less blur that benefited from precise structural constraints. For the larger case with 2048 image size in Fig.~\ref{fig:high_res3}, our method can still get more consistent result compared with LaMa.

\section{Limitations}

\begin{figure*}
\begin{centering}
\includegraphics[width=0.99\linewidth]{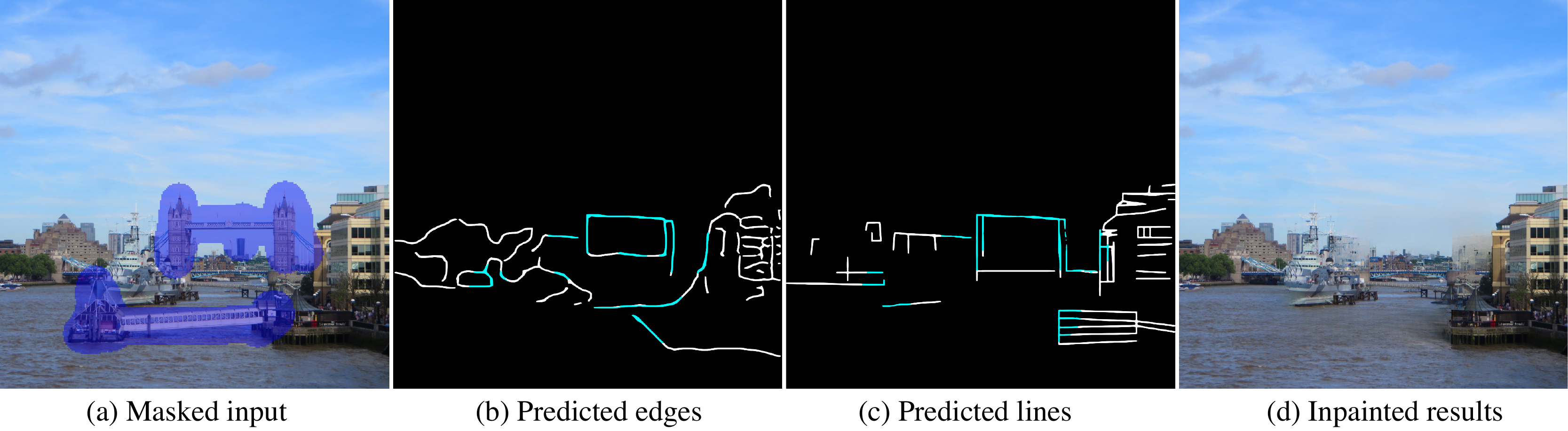}
\par\end{centering}
\caption{Failed 1024$\times$1024 results of our method. Some distant views failed to be described correctly by our grayscale sketch space of edges and lines. So these distant views are blurry.}
\label{fig:failed_case}
\end{figure*}

We summarize the limitation of our method in this section. As shown in Fig.~\ref{fig:failed_case}, since our method only recovers edges and lines in 256$\times$256, some distant views failed to be described correctly with the limited size. Therefore, some complex urban distant scenes can not be enhanced by the structures of canny edges and wireframe lines.

{\small
\bibliographystyle{ieee_fullname}
\bibliography{egbib}
}

\end{document}